\documentclass[a4paper,fleqn]{cas-dc}

\usepackage[authoryear]{natbib}

\usepackage{placeins}
\usepackage{booktabs}
\usepackage{kotex}
\usepackage{listings}
\usepackage{makecell}
\usepackage{bm}
\usepackage{soul}
\usepackage{tcolorbox}
\usepackage{lastpage}

\AtBeginDocument{%
}

\lstdefinestyle{promptstyle}{
  basicstyle=\ttfamily\scriptsize,
  backgroundcolor=\color{gray!4},
  frame=single,
  framerule=0.3pt,
  rulecolor=\color{gray!45},
  xleftmargin=0pt,
  xrightmargin=0pt,
  framexleftmargin=4pt,
  framexrightmargin=4pt,
  framextopmargin=3pt,
  framexbottommargin=3pt,
  breaklines=true,
  breakatwhitespace=false,
  columns=fullflexible,
  keepspaces=true,
  showstringspaces=false,
  aboveskip=3pt,
  belowskip=5pt
}

\lstdefinestyle{rolloutstyle}{
  basicstyle=\ttfamily\scriptsize,
  backgroundcolor=\color{gray!4},
  frame=single,
  framerule=0.3pt,
  rulecolor=\color{gray!45},
  xleftmargin=0pt,
  xrightmargin=0pt,
  framexleftmargin=4pt,
  framexrightmargin=4pt,
  framextopmargin=3pt,
  framexbottommargin=3pt,
  breaklines=true,
  breakatwhitespace=false,
  columns=fullflexible,
  keepspaces=true,
  showstringspaces=false,
  aboveskip=3pt,
  belowskip=6pt
}

\newcommand{\rollouttitle}[1]{%
  \vspace{0.45em}
  \noindent\textbf{\footnotesize #1}\par\vspace{0.15em}
}

\definecolor{qualwrong}{HTML}{C62828}
\definecolor{qualcorrect}{HTML}{2E7D32}

\newcommand{\qualwrong}[1]{%
    \textcolor{qualwrong}{\bfseries #1}%
}

\newcommand{\qualcorrect}[1]{%
    \textcolor{qualcorrect}{\bfseries #1}%
}

\newcommand{\qualwrongmath}[1]{%
    \textcolor{qualwrong}{\bm{#1}}%
}

\newcommand{\qualcorrectmath}[1]{%
    \textcolor{qualcorrect}{\bm{#1}}%
}

\newtcolorbox{qualresponsebox}[1]{
    colback=white,
    colframe=black!80,
    colbacktitle=black!8,
    coltitle=black,
    fonttitle=\bfseries,
    title={#1},
    boxrule=0.45pt,
    titlerule=0.4pt,
    arc=1.2mm,
    auto outer arc,
    boxsep=0pt,
    left=2.5mm,
    right=2.5mm,
    top=2mm,
    bottom=2mm,
    toptitle=1mm,
    bottomtitle=1mm,
    before skip=1.5mm,
    after skip=1.5mm
}

\tcbuselibrary{listings}

\def\tsc#1{\csdef{#1}{\textsc{\lowercase{#1}}\xspace}}
\tsc{WGM}
\tsc{QE}

\begin{document}
\let\WriteBookmarks\relax
\def\floatpagepagefraction{1}
\def\textpagefraction{.001}

\shorttitle{Beyond On-Policy Exploration: Integrating External Policy Rollouts for RL in dLLMs}    

\shortauthors{W.~Lee, J.~Kim, J.~Ko~and~W.~Rhee}  

\title [mode = title]{Beyond On-Policy Exploration: Integrating External Policy Rollouts for Reinforcement Learning in Diffusion Language Models}  



\author[1]{Wonseok~Lee}[orcid=0009-0004-8995-339X]
\ead{dnjstjr1017@snu.ac.kr}
\author[2]{Jimyeong~Kim}[orcid=0000-0003-2889-0861]
\ead{wlaud1001@snu.ac.kr}
\author[1]{Jungmin~Ko}[orcid=0009-0005-2009-6702]
\ead{jungminko@snu.ac.kr}
\author[1,3]{Wonjong~Rhee}[orcid=0000-0002-2590-8774]
\cormark[1]
\ead{wrhee@snu.ac.kr}

\affiliation[1]{organization={Interdisciplinary Program in Artificial Intelligence, Seoul National University},
            addressline={1 Gwanak-ro, Gwanak-gu}, 
            city={Seoul},
            postcode={08826}, 
            country={South Korea}}
\affiliation[2]{organization={Artificial Intelligence Institute, Seoul National University},
            addressline={1 Gwanak-ro, Gwanak-gu}, 
            city={Seoul},
            postcode={08826}, 
            country={South Korea}}
\affiliation[3]{organization={Department of Intelligence and Information, Seoul National University},
            addressline={1 Gwanak-ro, Gwanak-gu}, 
            city={Seoul},
            postcode={08826}, 
            country={South Korea}}

\cortext[1]{Corresponding author}



\begin{abstract}
Recent reinforcement learning methods for diffusion large language models (dLLMs) commonly rely on on-policy rollouts generated by the target dLLM itself. When successful on-policy rollouts are scarce, however, on-policy training may receive little positive reward and make only limited progress.
To mitigate this problem, we explore incorporating higher-reward rollouts generated by a stronger external policy alongside on-policy rollouts from the target dLLM.
However, directly incorporating these external rollouts introduces two practical challenges: differences in rollout length and instability when jointly processing rewards from on-policy and external rollouts.
To address these challenges, we propose External Rollout Integration with Length Control and Source-Specific Processing (ERILS), which controls external-rollout length and processes the rewards of on-policy and external rollouts separately.
Experiments on Sudoku, Countdown, and MATH500 under zero-shot evaluation show that ERILS improves multi-sample performance across all three tasks, with the largest gains on Sudoku.
On Sudoku, ERILS achieves 98.4\% best-of-4 completion accuracy, compared with 40.3\% for the strongest baseline.
ERILS also maintains approximately 90\% deterministic single-completion accuracy on Sudoku across generation lengths of 128, 256, and 512 tokens.
Our component analysis further shows that length-controlled external rollouts are more effective than uncontrolled external rollouts, and that source-specific reward processing avoids the training collapse observed with joint reward processing.
These results show that rollout construction and reward processing are important design dimensions when integrating external rollouts into dLLM reinforcement learning.
\end{abstract}





\maketitle

\section{Introduction}
\label{sec:intro}

\begin{figure*}[t]
\centering
\includegraphics[width=\textwidth]{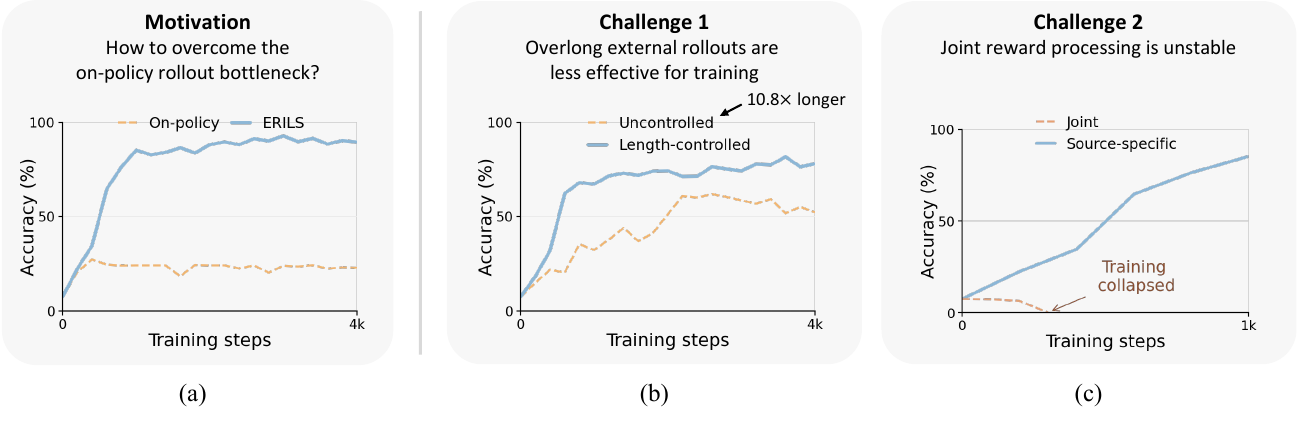}
\caption{
Motivation and challenges of integrating external rollouts into dLLM reinforcement learning, illustrated by training experiments on Sudoku.
(a) When successful on-policy rollouts are scarce, on-policy training remains at low accuracy, whereas ERILS substantially improves performance.
(b) External rollouts without length control are overlong and less effective for target-dLLM training than length-controlled external rollouts.
(c) Joint reward processing for on-policy and external rollouts leads to training collapse, whereas source-specific reward processing remains stable.
Together, these results motivate the use of rollout length control and source-specific reward processing when integrating external rollouts.
}
\label{fig:motivation_challenges}
\end{figure*}

Diffusion large language models (dLLMs) have recently emerged as non-autoregressive alternatives to autoregressive (AR) large language models (LLMs)~\citep{sahoo2024simple,lou2024discrete,shi2024simplified,gong2025scaling,nie2026large,zhu2026llada,ye2025dream}.
Unlike AR models, which generate tokens one by one from left to right, dLLMs start from a masked sequence and generate text by iteratively unmasking token positions across the sequence.
By predicting multiple token positions in parallel and conditioning on bidirectional context, dLLMs offer a distinct generation paradigm that is increasingly being explored for instruction-following and reasoning tasks.

To improve the reasoning capabilities of dLLMs, recent studies have applied reinforcement learning (RL) as a post-training approach~\citep{zhao2026d1,tang2026wd1,wang2026spg}.
Current dLLM RL methods commonly adopt reinforcement learning with verifiable rewards (RLVR), where the model generates rollouts, a task-specific verifier assigns each rollout a scalar reward, and policy optimization updates the model toward producing high-reward outputs~\citep{lambert2024tulu,shao2024deepseekmath,guo2025deepseek}.
Many of these methods follow a group-based optimization setting, related to Group Relative Policy Optimization (GRPO)~\citep{shao2024deepseekmath}, in which multiple rollouts are sampled for each prompt and their rewards are compared within the rollout group.

Applying RLVR to dLLMs differs from applying it to AR models in how sequence log-likelihoods are computed.
For AR language models, the sequence log-likelihood required for policy optimization can be computed exactly from token-level conditional probabilities under a causal attention mask.
For dLLMs, however, the corresponding exact sequence log-likelihood is generally intractable under the denoising formulation.
Existing dLLM RL methods have therefore focused on developing diffusion-compatible estimators or surrogate objectives, including one-step likelihood estimators~\citep{zhao2026d1}, weighted denoising objectives~\citep{tang2026wd1}, and variational lower- or upper-bound surrogates~\citep{wang2026spg}.

Although these methods make policy optimization feasible for dLLMs, they do not address a separate bottleneck in on-policy RLVR: whether the model can generate successful rollouts in the first place.
This bottleneck is particularly relevant in the on-policy setting commonly adopted by existing dLLM RL methods, where rollouts are sampled from the same dLLM policy being optimized~\citep{zhao2026d1,tang2026wd1,wang2026spg}.
When the current dLLM struggles to solve a task, it may rarely produce successful rollouts and thus receive little or no positive reward.
In such cases, improving the diffusion-compatible estimator alone may not compensate for the scarcity of positive-reward rollouts, and on-policy training may consequently make only limited progress, as illustrated in Fig.~\ref{fig:motivation_challenges}(a).

To mitigate this problem, we explore augmenting dLLM RL with higher-reward rollouts generated by a stronger external policy, providing positive reward even when the target dLLM rarely produces successful on-policy rollouts.
In this work, we instantiate the external policy with a stronger autoregressive language model and refer to its outputs as external rollouts.
We construct mixed rollout groups that contain both on-policy rollouts from the target dLLM and external rollouts from the stronger policy, following the broader idea of combining stronger-policy guidance with current-policy exploration in AR LLM reinforcement learning~\citep{yan2026learning}.
However, naïvely incorporating these external rollouts into the existing dLLM RL pipeline introduces two practical challenges.

The first challenge is that external rollouts yield only limited performance gains when they are directly used for dLLM RL, even though they receive substantially higher verifier rewards than the target dLLM's own samples.
To understand this discrepancy, we inspect the external rollouts and find that their completion lengths often substantially exceed the generation length used for on-policy rollout generation in existing dLLM RL methods~\citep{zhao2026d1,tang2026wd1,wang2026spg}.
This observation suggests that external rollout length should be taken into account when integrating external rollouts into dLLM RL.
We therefore introduce Rollout Length Control, which constructs external rollouts whose completion lengths are closer to the generation length while preserving their verifier-assessed quality.
As shown in Fig.~\ref{fig:motivation_challenges}(b), this substantially improves accuracy compared with using uncontrolled external rollouts.

The second challenge is that jointly processing the rewards from on-policy and external rollouts as a single group can destabilize training.
When on-policy and external rollouts for the same prompt are combined into a single rollout group, a naïve reward-processing approach is to compute one mean reward over all rollouts and process both sources relative to that mean.
In our experiments, this joint reward processing destabilizes optimization and eventually leads to training collapse, as shown in Fig.~\ref{fig:motivation_challenges}(c).
We therefore introduce Source-Specific Reward Processing, which applies separate reward-processing rules to on-policy and external rollouts.

We combine these two components into External Rollout Integration with Length Control and Source-Specific Processing (ERILS), a unified framework for integrating external rollouts into dLLM RL.
ERILS forms each rollout group with both on-policy rollouts from the target dLLM and external rollouts from a fixed external policy, while addressing the rollout-length and reward-processing problems identified in the naïve integration of these two sources.

We evaluate ERILS on Sudoku~\citep{arel_sudoku_2025}, Countdown~\citep{tinyzero}, and MATH500~\citep{lightman2024let} under zero-shot evaluation.
In stochastic multi-sample evaluation, ERILS improves performance across tasks.
On Sudoku, all ERILS best-of-$k$ completion accuracy scores exceed 90\%, whereas the strongest reproduced on-policy baseline reaches 40.3\% at best-of-4.
On Countdown, ERILS reaches 87.5\% Pass@4, compared with 76.6\% for SPG, and on MATH500 it achieves the highest results across all Pass@$k$ metrics among the compared methods.
Deterministic single-completion evaluation shows the most pronounced gain on Sudoku, where the evaluated on-policy dLLM RL baselines remain below 30\% accuracy, while ERILS exceeds 89\% accuracy across generation lengths of 128, 256, and 512 tokens.
Our component analysis shows that rollout length control makes external rollouts more effective for target-dLLM training, while source-specific reward processing avoids the instability observed under joint reward processing.
Together, these results demonstrate that external rollouts can be effectively integrated into dLLM RL when rollout construction and reward processing are handled explicitly.

Our main contributions are summarized as follows:
\begin{itemize}
\item We study the integration of external rollouts into dLLM reinforcement learning and identify two practical challenges concerning external rollout construction and reward processing for on-policy and external rollouts.
\item We propose External Rollout Integration with Length Control and Source-Specific Processing (ERILS), which combines rollout length control with source-specific reward processing to incorporate external rollouts alongside on-policy rollouts.
\item We analyze these two design choices and evaluate ERILS on Sudoku, Countdown, and MATH500 under both deterministic and multi-sample generation settings.
\end{itemize}

\section{Related Work}
\label{sec:related_work}

\subsection{Diffusion Language Models}

Diffusion language models generate text through iterative denoising rather than autoregressive next-token prediction.
Early discrete diffusion language models developed probabilistic frameworks for modeling text as categorical data, while subsequent masked diffusion formulations developed simple and effective denoising objectives for language modeling~\citep{lou2024discrete,sahoo2024simple,shi2024simplified}.
Building on these foundations, recent models such as DiffuLLaMA, LLaDA, LLaDA~1.5, and Dream have scaled Diffusion Language Models into large instruction-following and reasoning models, making diffusion language modeling an increasingly active research area~\citep{gong2025scaling,nie2026large,zhu2026llada,ye2025dream,nie2025scaling}.

Beyond this progress toward general-purpose language modeling, recent research has explored a broad range of architectural designs, generation mechanisms, and task-specific capabilities.
Several studies improve inference efficiency through faster denoising, caching, adaptive token updates, or confidence-aware decoding~\citep{wu2025fast,ma2025dkv,liu2025dllm,hu2025accelerating}.
Other work extends diffusion language modeling to long-context generation and develops blockwise or hybrid autoregressive--diffusion architectures that interpolate between parallel and sequential generation~\citep{liu2026longllada,arriola2025block,sun2025blockwise,wang2025diffusion}.
The bidirectional and non-causal structure of dLLMs has also motivated research on complex reasoning and planning, code generation, and diffusion-specific reasoning processes~\citep{ye2024diffusion,gong2026diffucoder,huang2026reinforcing}.
Collectively, these studies show that diffusion language modeling has developed into a broad research direction spanning model scaling, efficient generation, architectural design, and task-specific applications.

\subsection{Reinforcement Learning for Diffusion Language Models}

Reinforcement learning with verifiable rewards (RLVR) has become an important post-training approach for improving language models on mathematical, logical, and planning tasks~\citep{lambert2024tulu,shao2024deepseekmath,guo2025deepseek}.
Recent dLLM RL studies have similarly explored RLVR-style optimization, but adapting it to dLLMs is nontrivial because diffusion language models do not support the same tractable left-to-right sequence likelihood used by autoregressive language model policies.
A major line of dLLM RL research therefore focuses on constructing diffusion-compatible policy-optimization objectives.
The d1 framework introduces diffu-GRPO, which adapts GRPO to dLLMs using a one-step estimator of sequence log-probability~\citep{zhao2026d1}.
In contrast, wd1 avoids explicit policy-ratio estimation and instead optimizes an advantage-weighted likelihood objective~\citep{tang2026wd1}.
SPG further reduces the bias associated with likelihood approximation by selecting lower- or upper-bound surrogates according to the sign of the group-relative advantage, together with blockwise masking for stable estimation~\citep{wang2026spg}.
More recent work has also explored more accurate likelihood approximation together with importance-sampling correction for dLLM policy optimization~\citep{zhao2025diffpo}.
Despite differences in their optimization formulations, these methods primarily focus on improving the policy update while relying on rollouts sampled from the current dLLM.
Consequently, their updates depend on the reasoning trajectories that the current dLLM can discover through its own sampling process.

\begin{figure*}[!t]
\centering
\includegraphics[width=\textwidth]{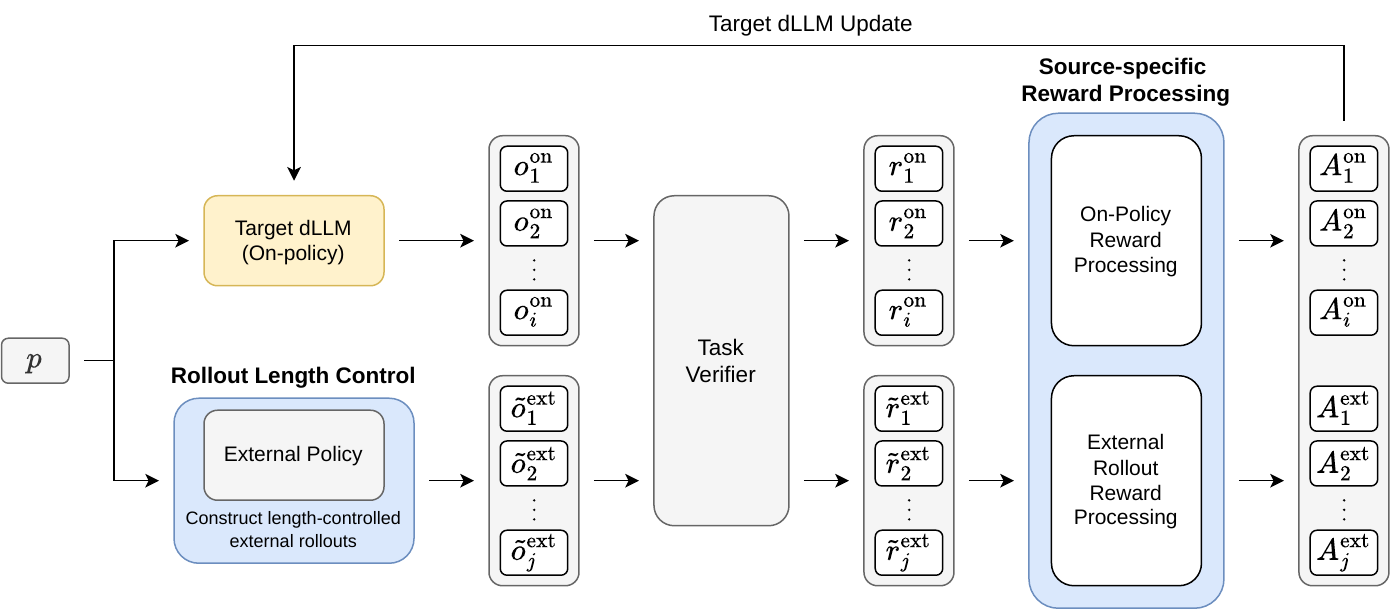}
\caption{
Overview of ERILS.
ERILS combines on-policy rollouts from the target dLLM with length-controlled external rollouts constructed using a fixed external policy.
Rollout Length Control produces external rollouts compatible with the target dLLM's generation length before they are scored by the task verifier.
ERILS then applies Source-Specific Reward Processing, which processes on-policy rewards and external-rollout rewards separately.
The resulting contributions are averaged over the mixed rollout group, and only the target dLLM is updated.
}
\label{fig:method_overview}
\end{figure*}

A related line of work addresses the limits of on-policy exploration by using reasoning traces from sources beyond the current policy.
For autoregressive LLMs, trajectories generated by a stronger policy have been combined with on-policy rollouts, with advantages computed over the mixed rollout group and regularized importance sampling used for off-policy updates~\citep{yan2026learning}.
Because this formulation relies on token-level policy ratios, it does not directly extend to dLLM reinforcement learning.
For dLLMs, prior work has instead augmented all-wrong on-policy rollout groups by conditioning generation on partial ground-truth reasoning traces through inpainting~\citep{zhao2026inpainting}.
This approach assumes access to a ground-truth reasoning trace for each training problem and uses selected parts of the trace as generation context.
Our setting does not condition the target dLLM on partial reasoning traces.
Instead, a separate stronger policy independently generates external rollouts, and we study their construction and reward processing when they are combined with on-policy dLLM rollouts.

\section{Preliminary}
\label{sec:preliminary}

\subsection{Reinforcement Learning for LLMs}
\label{sec:prelim_rl_llms}

Let $R(p,o)$ denote the scalar verifier reward assigned to completion $o$ for prompt $p$.
Given a policy $\pi_\theta$ with parameters $\theta$ and a prompt distribution $\mathcal D$, the reinforcement-learning objective is to maximize the expected reward of generated completions:
\begin{equation}
J(\theta) = \mathbb E_{ p\sim\mathcal D, o\sim\pi_\theta(\cdot\mid p) } \left[ R(p,o) \right].
\label{eq:rl_objective}
\end{equation}
The gradient of this objective can be written as
\begin{equation}
\nabla_\theta J(\theta) = \mathbb E_{ p\sim\mathcal D, o\sim\pi_\theta(\cdot\mid p) } \left[
R(p,o) \nabla_\theta \log\pi_\theta(o\mid p)
\right],
\label{eq:pg_objective}
\end{equation}

Group-based RL methods~\citep{shao2024deepseekmath,liu2025understanding} construct group-relative advantages from multiple completions generated for the same prompt.
For each prompt $p\sim\mathcal D$, let $o_1,\ldots,o_G$ be independently
sampled from $\pi_\theta(\cdot\mid p)$, with rewards $r_i=R(p,o_i)$.
We use mean-centered rewards~\citep{liu2025understanding} to define the group-relative advantages:
\begin{equation}
A_i = r_i - \bar r,
\label{eq:group_adv}
\end{equation}
where $\bar r$ denotes the mean reward of the $G$ completions in the group.
The corresponding group-based policy update takes the form
\begin{equation}
\mathbf g_{\mathrm{group}}(\theta) = \mathbb E_{p,\{o_i\}_{i=1}^{G}} \left[\frac{1}{G}\sum_{i=1}^{G}A_i\nabla_\theta \log \pi_\theta(o_i\mid p)\right].
\label{eq:grpo_pg}
\end{equation}

\subsection{Reinforcement Learning for dLLMs}

A masked dLLM generates a completion by iteratively denoising an initial sequence of $L_{\mathrm{gen}}$ masked positions.
We refer to $L_{\mathrm{gen}}$ as the generation length, which denotes the number of completion positions allocated at the start of generation rather than the realized completion length.

Unlike autoregressive generation, iterative denoising in a masked dLLM does not provide a tractable left-to-right factorization of the sequence likelihood.
Consequently, the sequence log-likelihood term $\log\pi_\theta(o_i\mid p)$ in Eq.~\ref{eq:grpo_pg} cannot be evaluated in the same manner as for an autoregressive policy.

To formulate group-relative optimization for masked dLLMs, let $\mathcal S_\theta(p,o_i)$ denote a diffusion-compatible surrogate objective for completion $o_i$.
Replacing the log-likelihood gradient in Eq.~\ref{eq:grpo_pg} with the gradient of this surrogate yields
\begin{equation}
\mathbf g_{\mathrm{dLLM}}(\theta) = \mathbb E_{p,\{o_i\}_{i=1}^{G}} \left[
\frac{1}{G} \sum_{i=1}^{G} A_i \nabla_\theta \mathcal S_\theta(p,o_i)
\right].
\label{eq:dllm_pg_objective}
\end{equation}
The specific form of $\mathcal S_\theta$ depends on the likelihood estimator or variational surrogate used for dLLM optimization~\citep{zhao2026d1,tang2026wd1,wang2026spg}.

\section{External Rollout Integration with Length Control and Source-Specific Processing}
\label{sec:method}

We consider reinforcement learning for a target dLLM $\pi_\theta$ with access to a fixed external policy.
For each prompt $p$, the target dLLM generates $G_{\mathrm{on}}$ on-policy rollouts:
\begin{equation}
    o_i^{\mathrm{on}} \sim \pi_\theta(\cdot\mid p),\qquad i=1,\ldots,G_{\mathrm{on}}
\end{equation}
The on-policy rollouts are evaluated using the task verifier:
\begin{equation}
r_i^{\mathrm{on}} = R(p,o_i^{\mathrm{on}}).
\end{equation}

ERILS uses the external policy through Rollout Length Control and combines the resulting external rollouts with the on-policy rollouts using Source-Specific Reward Processing, as illustrated in Fig.~\ref{fig:method_overview}.
We describe Rollout Length Control in Section~\ref{sec:length_control} and Source-Specific Reward Processing in Section~\ref{sec:mixed_learning}.

\subsection{Rollout Length Control}
\label{sec:length_control}

We found that the external AR policy often generates rollouts substantially longer than the fixed generation length used for on-policy dLLM rollouts in prior dLLM RL methods, typically $L_{\mathrm{gen}}=256$~\citep{zhao2026d1,tang2026wd1,wang2026spg}.
In ERILS, we therefore use $L_{\mathrm{gen}}$ as a practical reference when constructing external rollouts with lengths comparable to the target dLLM’s generation length.

By default, we generate length-controlled external completions by appending a length-control instruction to the task prompt used for on-policy rollout generation.
The instruction encourages the external policy to generate a completion with a length close to $L_{\mathrm{gen}}$.
For tasks where this procedure does not reliably produce completions with comparable lengths, we use a two-stage rewriting procedure.
The external policy first generates a completion from the original task prompt and then rewrites it into a more compact form with a length closer to $L_{\mathrm{gen}}$.
The rewrite is instructed to preserve the final answer while retaining only the essential reasoning from the first-stage completion.

Using this procedure, we construct $G_{\mathrm{ext}}$ length-controlled external rollouts with the external policy for each prompt.
We denote the resulting external rollout by $\widetilde{o}_j^{\mathrm{ext}}$, $j=1,\ldots,G_{\mathrm{ext}}$ with rewards
\begin{equation}
\widetilde{r}_j^{\mathrm{ext}} = R(p,\widetilde{o}_j^{\mathrm{ext}}).
\end{equation}
Together with the $G_{\mathrm{on}}$ on-policy rollouts defined above, these length-controlled external rollouts form the mixed rollout group used for optimization.
The exact instructions and additional implementation details are provided in Appendix~\ref{sec:appendix_rollout_construction}.

\subsection{Mixed-Rollout Optimization with Source-Specific Reward Processing}
\label{sec:mixed_learning}

On-policy and external rollouts are generated by different policies and can exhibit substantially different reward distributions.
When external rollouts typically receive higher rewards than on-policy rollouts, centering all rewards using a single mean over the mixed rollout group can assign negative advantages to many on-policy rollouts, even when they receive relatively high rewards among the target dLLM's own samples.
ERILS avoids this cross-source coupling by processing the rewards from the two rollout sources separately.
For on-policy rollouts, we use the mean-centered group-relative advantage.
For external rollouts, we directly use the verifier reward assigned to the final length-controlled rollout:
\begin{equation}
\begin{aligned}
A_i^{\mathrm{on}} &= r_i^{\mathrm{on}} - \frac{1}{G_{\mathrm{on}}} \sum_{\ell=1}^{G_{\mathrm{on}}} r_\ell^{\mathrm{on}},\\
A_j^{\mathrm{ext}} &= \widetilde{r}_j^{\mathrm{ext}}.
\end{aligned}
\label{eq:source_specific_reward_processing}
\end{equation}
Accordingly, the on-policy advantages depend only on rewards within the on-policy group.

Let $\mathcal S_\theta(p,o)$ denote a diffusion-compatible surrogate objective for completion $o$.
Using Eq.~\ref{eq:source_specific_reward_processing}, the target dLLM is updated with the following mixed-rollout gradient estimator:
\begin{equation}
\begin{aligned}
\mathbf g_{\mathrm{ERILS}}(\theta) = \frac{1}{G_{\mathrm{on}}+G_{\mathrm{ext}}}
&\bigg(
\sum_i A_i^{\mathrm{on}} \nabla_\theta \mathcal S_\theta (p,o_i^{\mathrm{on}})\\
+ \sum_j A_j^{\mathrm{ext}} &\nabla_\theta \mathcal S_\theta (p,\widetilde{o}_j^{\mathrm{ext}})
\bigg)
\end{aligned}
\label{eq:erils_update}
\end{equation}

\section{Experiments}
\label{sec:experiments}

\subsection{Experimental Setup}
\label{sec:exp_setup}

\paragraph*{Tasks.}
We conduct experiments on Sudoku~\citep{arel_sudoku_2025}, Countdown~\citep{tinyzero}, and MATH500~\citep{lightman2024let}.
Following d1 and wd1, we use zero-shot prompts without in-context demonstrations for all three tasks~\citep{zhao2026d1,tang2026wd1}.

\paragraph*{Models and rollout configuration.}
Following prior dLLM RL studies~\citep{zhao2026d1,tang2026wd1,wang2026spg}, we use LLaDA-8B-Instruct~\citep{nie2026large} as the target dLLM $\pi_\theta$.
On-policy rollouts are generated with $L_{\mathrm{gen}}=256$ during training, following prior dLLM RL settings~\citep{zhao2026d1,wang2026spg}.
External rollouts are generated using Qwen3-30B-A3B-Instruct-2507~\citep{qwen3technicalreport}, which remains fixed throughout training.
External rollouts are generated using the Rollout Length Control described in Section~\ref{sec:length_control}.
For Countdown and MATH500, we append a length-control instruction to the task prompt used for on-policy rollout generation and directly generate the final completion.
For Sudoku, because this approach does not reliably produce completions with lengths comparable to $L_{\mathrm{gen}}$, we instead use the two-stage rewriting procedure described in Section~\ref{sec:length_control}, which rewrites an initially generated completion into a more compact form while preserving its final answer.
The selected procedure is applied uniformly to all prompts within each task.
The default ERILS configuration uses $G_{\mathrm{on}}=4$ on-policy rollouts and $G_{\mathrm{ext}}=2$ external rollouts for each prompt.

\paragraph*{Training configuration.}
For each task, we fine-tune the target dLLM using Low-Rank Adaptation (LoRA) with rank $r=128$ and scaling factor $\alpha=64$.
We instantiate the diffusion-compatible surrogate objective $\mathcal S_\theta(p,o)$ using SPG~\citep{wang2026spg} for both rollout sources and otherwise follow the SPG generation and optimization settings.
Additional training and implementation details, including the number of training steps used for each task and analysis experiment, are provided in Appendix~\ref{sec:appendix_experimental_details}.

\paragraph*{Baselines.}
We compare ERILS with representative on-policy dLLM RL methods, including d1, wd1, and SPG~\citep{zhao2026d1,tang2026wd1,wang2026spg}.
Because the original SPG paper evaluates Sudoku using three solved demonstrations~\citep{wang2026spg}, we reproduce SPG under the same zero-shot protocol used for ERILS.
Additional details are provided in Appendix~\ref{sec:appendix_baselines}.

\begin{table*}[t]
\centering
\caption{
Multi-sample zero-shot evaluation on Sudoku, Countdown, and MATH500 with temperature 0.9.
Sudoku reports best-of-$k$ completion accuracy, where completion accuracy is computed over the originally empty cells.
Countdown and MATH500 baseline results are taken from SPG~\citep{wang2026spg}.
All Sudoku results are obtained from our evaluations.
Bold and underline indicate the best and second-best results, respectively.
}
\label{tab:multi_sample_evaluation}
\small
\setlength{\tabcolsep}{4pt}
\begin{tabular}{l|cccc|cccc|cccc}
\toprule
& \multicolumn{4}{c|}{\textbf{Sudoku (Best-of-$k$)}}
& \multicolumn{4}{c|}{\textbf{Countdown (Pass@$k$)}}
& \multicolumn{4}{c}{\textbf{MATH500 (Pass@$k$)}} \\
\cmidrule(lr){2-5}
\cmidrule(lr){6-9}
\cmidrule(lr){10-13}

\textbf{Model}
& $k=1$
& $k=2$
& $k=3$
& $k=4$
& $k=1$
& $k=2$
& $k=3$
& $k=4$
& $k=1$
& $k=2$
& $k=3$
& $k=4$ \\
\midrule

LLaDA-8B-Instruct~\citep{nie2026large}
& 9.2 & 14.5 & 17.5 & 20.6
& 15.8 & 28.1 & 37.7 & 45.3
& 31.5 & 40.9 & 45.7 & 48.8 \\

LLaDA-1.5~\citep{zhu2026llada}
& 11.4 & 17.4 & 20.7 & 23.3
& 18.2 & 32.1 & 42.5 & 50.0
& 32.6 & 42.2 & 47.4 & 50.4 \\

d1~\citep{zhao2026d1}
& - & - & - & -
& 24.5 & 40.4 & 51.4 & 60.6
& 34.3 & 43.1 & 48.0 & 52.0 \\

wd1~\citep{tang2026wd1}
& - & - & - & -
& 44.3 & 60.6 & 68.0 & 73.1
& 36.0 & 44.9 & 49.9 & 53.6 \\

UniGRPO~\citep{wang2026spg}
& - & - & - & -
& 36.8 & 55.2 & 65.0 & 72.3
& 34.7 & 43.9 & 49.5 & 53.2 \\

SPG~\citep{wang2026spg}
& \underline{25.2}
& \underline{33.3}
& \underline{37.3}
& \underline{40.3}
& \textbf{67.5}
& \underline{72.5}
& \underline{75.1}
& \underline{76.6}
& \underline{36.5}
& \underline{46.0}
& \underline{51.2}
& \underline{55.6} \\

\midrule

ERILS
& \textbf{91.2}
& \textbf{95.4}
& \textbf{97.1}
& \textbf{98.4}
& \underline{67.2}
& \textbf{78.9}
& \textbf{84.8}
& \textbf{87.5}
& \textbf{37.8}
& \textbf{47.0}
& \textbf{52.0}
& \textbf{56.0} \\

\bottomrule
\end{tabular}
\end{table*}
\begin{table*}[t]
\centering
\small
\setlength{\tabcolsep}{5pt}
\caption{
Deterministic single-completion zero-shot evaluation on Sudoku, Countdown, and MATH500 at $L_{\mathrm{gen}}\in\{128,256,512\}$ using temperature 0.0.
The $\dagger$ symbol marks our zero-shot reproduction of SPG on Sudoku.
Bold values indicate the best result in each column.
Missing entries indicate unavailable or non-comparable evaluations.
}
\begin{tabular}{l|ccc|ccc|ccc}
\toprule
& \multicolumn{3}{c|}{\textbf{Sudoku}}
& \multicolumn{3}{c|}{\textbf{Countdown}}
& \multicolumn{3}{c}{\textbf{MATH500}} \\
\cmidrule(lr){2-4}
\cmidrule(lr){5-7}
\cmidrule(lr){8-10}
\textbf{Method}
& 128 & 256 & 512
& 128 & 256 & 512
& 128 & 256 & 512 \\
\midrule
LLaDA-8B-Instruct~\citep{nie2026large}
& 11.7 & 6.7 & 5.5 
& 18.8 & 16.8 & 16.8
& 28.2 & 32.4 & 34.6 \\

LLaDA-1.5~\citep{zhu2026llada}
& - & - & - 
& 21.9 & 21.1 & 21.5
& 26.8 & 32.2 & 35.8 \\

d1~\citep{zhao2026d1}
& 22.1 & 16.7 & 9.5 
& 34.8 & 32.0 & 42.2
& 33.8 & 38.6 & 40.2 \\

wd1~\citep{tang2026wd1}
& - & 25.2 & 24.2
& - & 51.2 & 46.1
& - & 34.4 & 39.0 \\

UniGRPO~\citep{wang2026spg}
& - & - & - 
& 44.5 & 43.0 & 57.0
& 32.4 & 37.4 & 39.4 \\

SPG~\citep{wang2026spg}
& 27.3$\dagger$ & 26.8$\dagger$ & 26.4$\dagger$
& 68.8 & 70.7 & 70.3
& 33.4 & 40.0 & \textbf{41.8} \\

\midrule

ERILS
& \textbf{91.7} & \textbf{91.2} & \textbf{89.5}
& \textbf{69.9} & \textbf{73.4} & \textbf{73.8}
& \textbf{34.8} & \textbf{40.2} & 39.4 \\
\bottomrule
\end{tabular}
\label{tab:main_benchmark}
\end{table*}

\paragraph*{Metrics.}
We evaluate models under two generation settings: multi-sample evaluation and deterministic single-completion evaluation.
Multi-sample evaluation generates multiple completions for the same prompt using stochastic decoding to assess whether the model can generate a correct solution through sampling, whereas deterministic single-completion evaluation generates one completion for each prompt using a fixed decoding rule.
For Countdown and MATH500, each completion receives a binary correctness score for the entire problem; we therefore report Pass@$k$ under multi-sample evaluation~\citep{chen2021evaluating,chen2026does,shao2024deepseekmath}, defined as the fraction of problems for which at least one of the $k$ sampled completions is correct, and accuracy under deterministic single-completion evaluation.
For Sudoku, each completion is evaluated using completion accuracy, defined as the fraction of originally empty cells filled with the correct digits.
Accordingly, we report best-of-$k$ completion accuracy under multi-sample evaluation by taking the highest completion accuracy among the $k$ sampled completions, and the completion accuracy of the single completion under deterministic evaluation.

\paragraph*{Evaluation protocol.}
We evaluate models using semi-autoregressive confidence-based decoding with block size 32 and $L_{\mathrm{gen}}/2$ diffusion steps~\citep{nie2026large}.
Multi-sample evaluation uses $L_{\mathrm{gen}}=256$, whereas deterministic single-completion evaluation uses $L_{\mathrm{gen}}\in\{128,256,512\}$.
Following the checkpoint reporting practice used in prior dLLM RL studies~\citep{zhao2026d1,wang2026spg}, 
we evaluate checkpoints from the main experiments at regular intervals and report the best observed result for each reported evaluation configuration.
For analysis experiments, we evaluate checkpoints every 100 steps on a fixed subset of 128 examples from the Sudoku evaluation set to track performance over training.
Additional evaluation details are provided in Appendix~\ref{sec:appendix_experimental_details}.

\subsection{Main Benchmark Results}
\label{sec:benchmark_results}

We compare ERILS with the baselines under the two evaluation settings defined above. Overall, ERILS shows the clearest advantage on Sudoku, where it substantially outperforms all baselines in both multi-sample and deterministic single-completion evaluation. On Countdown and MATH500, ERILS also improves multi-sample performance and remains competitive under deterministic generation.

\paragraph*{Multi-sample evaluation.}
Table~\ref{tab:multi_sample_evaluation} reports multi-sample evaluation results.
The largest improvement is observed on Sudoku, where ERILS improves best-of-$k$ completion accuracy from 25.2\% to 91.2\% at $k=1$ and from 40.3\% to 98.4\% at $k=4$ compared with SPG. On Countdown, ERILS performs comparably to SPG at $k=1$ but achieves a 6.4 percentage-point improvement at $k=2$, with the advantage increasing for larger $k$. On MATH500, ERILS consistently outperforms the compared methods across all evaluated values of $k$.

\begin{figure}[pos=t]
\centering
\includegraphics[width=\columnwidth]{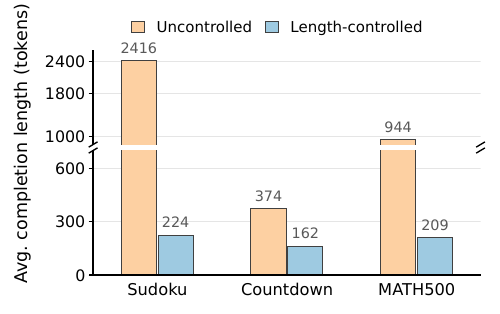}
\caption{
Average completion lengths of uncontrolled and length-controlled external rollouts.
Rollout Length Control substantially reduces overlong external completions and produces rollout lengths compatible with the on-policy generation length $L_{\mathrm{gen}}=256$.
}
\label{fig:external_rollout_length}
\end{figure}

\paragraph*{Deterministic single-completion evaluation.}
Table~\ref{tab:main_benchmark} reports deterministic single-completion results. ERILS achieves substantial gains on Sudoku across all generation lengths, with completion accuracy consistently remaining around 90\%, while the strongest baseline remains below 30\%. This demonstrates that the Sudoku gains are not limited to the multi-sample setting. On Countdown, ERILS outperforms SPG across all three generation lengths. On MATH500, ERILS outperforms SPG at $L_{\mathrm{gen}}\in\{128,256\}$, while achieving slightly lower accuracy at $L_{\mathrm{gen}}=512$.  Overall, ERILS achieves the largest gains on Sudoku and demonstrates competitive performance on Countdown and MATH500 across different generation lengths.

\subsection{Component Analysis}

In this section, we analyze the two key components of ERILS: length-controlled external rollouts and source-specific reward processing.

\subsubsection{Effect of Length-Controlled External Rollouts}
\label{sec:analysis_rollout_construction}

We examine the role of Rollout Length Control in making external rollouts more suitable for target-dLLM training. We first analyze the lengths of external rollouts with and without length control, and then compare their effectiveness for target-dLLM training.

\begin{figure}[pos=t]
\centering
\includegraphics[width=\columnwidth]{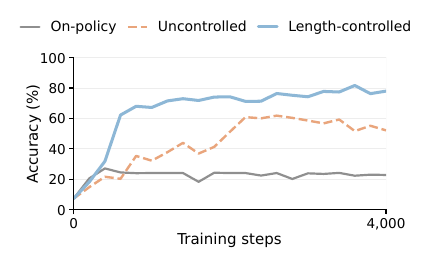}
\caption{
Effect of rollout length control on Sudoku training.
Uncontrolled external rollouts improve over on-policy-only training, while length-controlled external rollouts provide a substantially larger improvement.
}
\label{fig:length_control_training}
\end{figure}

\begin{figure}[pos=t]
\centering
\includegraphics[width=\columnwidth]{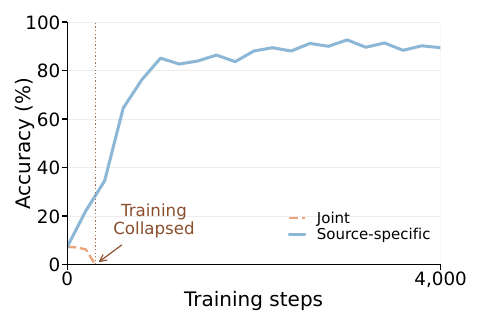}
\caption{
Effect of reward processing on mixed-rollout training for Sudoku.
"Joint" and "Source-specific" denote joint reward processing and source-specific reward processing, respectively.
Both configurations use four on-policy rollouts and two length-controlled external rollouts per prompt.
Joint reward processing leads to training collapse, whereas source-specific reward processing remains stable and substantially improves accuracy.
}
\label{fig:mix_controlled_training}
\end{figure}

Figure~\ref{fig:external_rollout_length} compares the average completion lengths of external rollouts generated with and without length control. Without length control, the external policy often produces completions substantially longer than $L_{\mathrm{gen}}=256$, reaching 2,416 tokens on Sudoku and 944 tokens on MATH500. After applying Rollout Length Control, the average completion lengths decrease to 224, 162, and 209 tokens on Sudoku, Countdown, and MATH500, respectively.

We next examine whether length control makes external rollouts more effective for target-dLLM training.
To isolate the effect of length control, we compare three configurations on Sudoku: external-only training with uncontrolled external rollouts, external-only training with length-controlled external rollouts, and on-policy-only training as a reference.
The two external-only configurations use the same optimization procedure and differ only in whether Rollout Length Control is applied.
The detailed analysis settings are provided in Appendix~\ref{sec:appendix_experimental_details}.
As shown in Fig.~\ref{fig:length_control_training}, both external-only configurations improve over on-policy-only training, but length-controlled external rollouts yield substantially larger and more sustained gains than uncontrolled external rollouts. This gap shows that Rollout Length Control plays an important role in making external rollouts effective for target-dLLM training.

\begin{figure}[pos=t]
\centering
\includegraphics[width=\columnwidth]{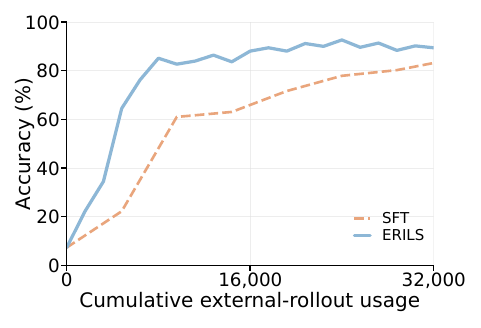}
\caption{
Comparison between ERILS and supervised fine-tuning on Sudoku with matched cumulative external-rollout usage.
}
\label{fig:erils_vs_sft_external_usage}
\end{figure}

\begin{figure}[pos=t]
\centering
\includegraphics[width=\columnwidth]{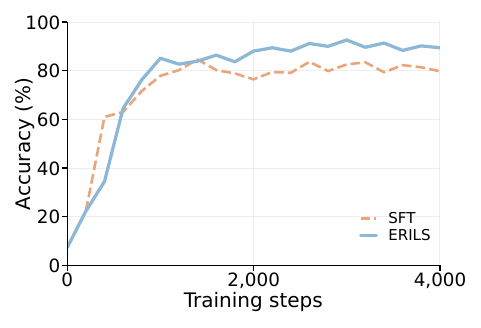}
\caption{
Comparison between ERILS and supervised fine-tuning on Sudoku with matched training steps.
}
\label{fig:erils_vs_sft_training_steps}
\end{figure}

\subsubsection{Effect of Source-Specific Reward Processing}
\label{sec:analysis_mixed_rollout}

We next examine the effectiveness of source-specific reward processing in mixed-rollout training.
Specifically, we compare two variants that differ only in how advantages are computed.
Joint reward processing treats all rollouts as a single group and centers the rewards of both sources using the mean reward over the mixed rollout group:
\begin{equation}
\begin{aligned}
A_i^{\mathrm{on,mix}} &= r_i^{\mathrm{on}} - \bar{r},\\
A_j^{\mathrm{ext,mix}} &= \widetilde{r}_j^{\mathrm{ext}} - \bar{r},
\end{aligned}
\label{eq:joint_advantage}
\end{equation}
where
\begin{equation}
\bar{r} = \frac
{\sum_{\ell=1}^{G_{\mathrm{on}}} r_\ell^{\mathrm{on}} + \sum_{\ell=1}^{G_{\mathrm{ext}}} \widetilde{r}_\ell^{\mathrm{ext}}}
{G_{\mathrm{on}}+G_{\mathrm{ext}}}.
\label{eq:joint_reward_mean}
\end{equation}
The source-specific variant instead processes the two rollout sources according to Eq.~\ref{eq:source_specific_reward_processing}.

As shown in Fig.~\ref{fig:mix_controlled_training}, the mixed-rollout variant with joint reward processing collapses during training, whereas the source-specific variant remains stable and achieves substantially higher accuracy. This result shows that processing rewards separately by source is important for stable and effective mixed-rollout training. We further analyze the collapse under joint reward processing in Section~\ref{sec:analysis_joint_processing_collapse}.

\begin{figure}[pos=t]
\centering
\includegraphics[width=\columnwidth]{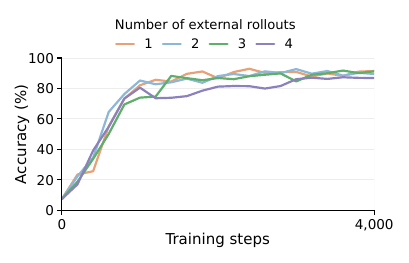}
\caption{
Sensitivity to rollout composition on Sudoku.
The total rollout-group size is fixed at six, while $G_{\mathrm{ext}}$ is varied from one to four and $G_{\mathrm{on}}=6-G_{\mathrm{ext}}$.
Increasing the number of external rollouts beyond two does not yield a consistent improvement.
}
\label{fig:external_rollout_count}
\end{figure}

\begin{figure}[pos=t]
\centering
\includegraphics[width=\columnwidth]{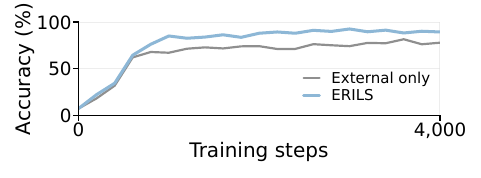}
\caption{
Comparison between ERILS and external-only training on Sudoku.
ERILS combines external and on-policy rollouts, whereas the external-only configuration updates the target dLLM using only external rollouts.
}
\label{fig:external_vs_mix}
\end{figure}

\subsection{Additional Evaluations}
\subsubsection{Comparison with Supervised Fine-Tuning}
\label{sec:sft_comparison}

Since ERILS leverages length-controlled rollouts from a stronger external policy, a natural question is whether its gains arise simply from exposure to these external rollouts or from how they are used during reinforcement learning.
We therefore compare ERILS with a supervised fine-tuning (SFT) baseline trained on length-controlled external rollouts.
ERILS combines external rollouts with on-policy rollouts generated during reinforcement learning, whereas SFT is trained only on external rollouts.
ERILS and SFT use the same target dLLM, LoRA configuration, training split, and GPU hardware configuration.
Additional implementation details for the SFT baseline are provided in Appendix~\ref{sec:appendix_sft}.

Because ERILS and SFT differ in both optimization procedure and exposure to external rollouts, a single budget axis is insufficient for a fair comparison.
We therefore compare them by cumulative external-rollout usage and training steps.
Here, external-rollout usage refers to the number of external prompt-completion pairs used during training.
Under this definition, each ERILS prompt group contributes two external-rollout uses, whereas each SFT prompt group contributes six.
Training steps track progress under each method's respective optimization procedure.
All curves follow the analysis protocol described in Section~\ref{sec:exp_setup}.

Figures~\ref{fig:erils_vs_sft_external_usage} and~\ref{fig:erils_vs_sft_training_steps} compare ERILS and SFT under matched cumulative external-rollout usage and matched training steps, respectively.
Under matched external-rollout usage, ERILS maintains a clear performance advantage over SFT, achieving a peak accuracy of 89.5\% compared with 83.2\% for SFT.
Under matched training steps, SFT improves more rapidly in early training, but ERILS eventually surpasses SFT and maintains higher accuracy throughout most of the later trajectory.
These results indicate that ERILS provides a more effective way to leverage length-controlled external rollouts than supervised fine-tuning.

\begin{figure*}[pos=t]
    \centering
    \includegraphics[width=\textwidth]{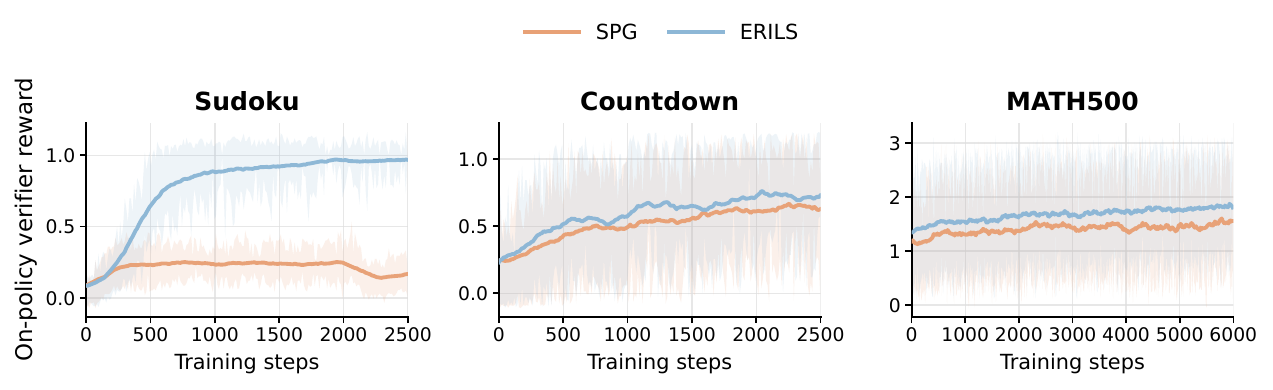}
    \caption{
    On-policy verifier rewards during ERILS and SPG training on Sudoku, Countdown, and MATH500.
    SPG is reproduced under the same zero-shot setup and training-prompt order used for ERILS.
    Solid lines show mean values smoothed using a centered moving average over 25 logged points, with statistics recorded every 12 training steps.
    Shaded regions indicate the standard deviation computed from the unsmoothed statistics.
    }
    \label{fig:erils_reward_dynamics}
\end{figure*}

\subsubsection{Sensitivity to Rollout Composition}
\label{sec:rollout_composition_sensitivity}

We examine the sensitivity of ERILS to rollout composition by varying the number of external rollouts in the mixed rollout group. Specifically, we vary $G_{\mathrm{ext}}\in\{1,2,3,4\}$ while keeping the total number of rollouts per group fixed at six, with $G_{\mathrm{on}}=6-G_{\mathrm{ext}}$.
As shown in Fig.~\ref{fig:external_rollout_count}, ERILS maintains comparable performance across all evaluated rollout compositions. We use $G_{\mathrm{ext}}=2$ and $G_{\mathrm{on}}=4$ as the default configuration.

\subsubsection{Effect of Combining External and On-Policy Rollouts}
\label{sec:external_vs_mix}

ERILS combines external rollouts with on-policy rollouts from the target dLLM.
A simpler alternative is to update the target dLLM using only external rollouts.
To examine whether the on-policy component provides an additional benefit, we compare ERILS with this external-only configuration under the same setup used in the component analysis.

As shown in Fig.~\ref{fig:external_vs_mix}, the external-only configuration reaches a peak accuracy of 81.6\%, whereas ERILS reaches 92.7\%.
This result shows that, in our setting, combining external and on-policy rollouts is more effective than training with external rollouts alone.
It also suggests that the improvement of ERILS is not explained solely by exposure to high-reward external rollouts; retaining on-policy rollouts provides an additional benefit.

\begin{figure}[pos=t]
\centering
\includegraphics[width=\columnwidth]{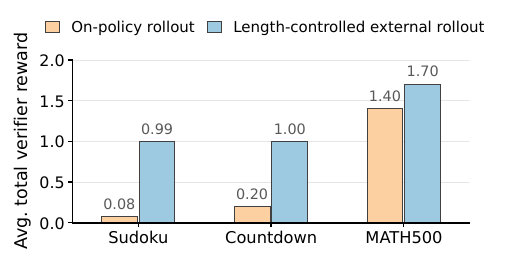}
\caption{
Average verifier rewards of initial on-policy rollouts and length-controlled external rollouts across the three tasks.
Rewards follow the task-specific verifier scales and are compared within each task.
}
\label{fig:rollout_source_reward}
\end{figure}

\subsubsection{Reward Dynamics}
\label{sec:reward_dynamics}

We compare how the on-policy verifier rewards evolve during ERILS and SPG training.
For this analysis, we reproduce SPG under the same zero-shot setup and training-prompt order used for ERILS.
As shown in Fig.~\ref{fig:erils_reward_dynamics}, ERILS achieves higher on-policy verifier rewards than SPG across all three tasks.

\section{Discussion}

\subsection{Verifier Rewards of External Rollouts}
\label{sec:external_rollout_reward}

\begin{figure*}[pos=t]
\centering
\includegraphics[width=\textwidth]{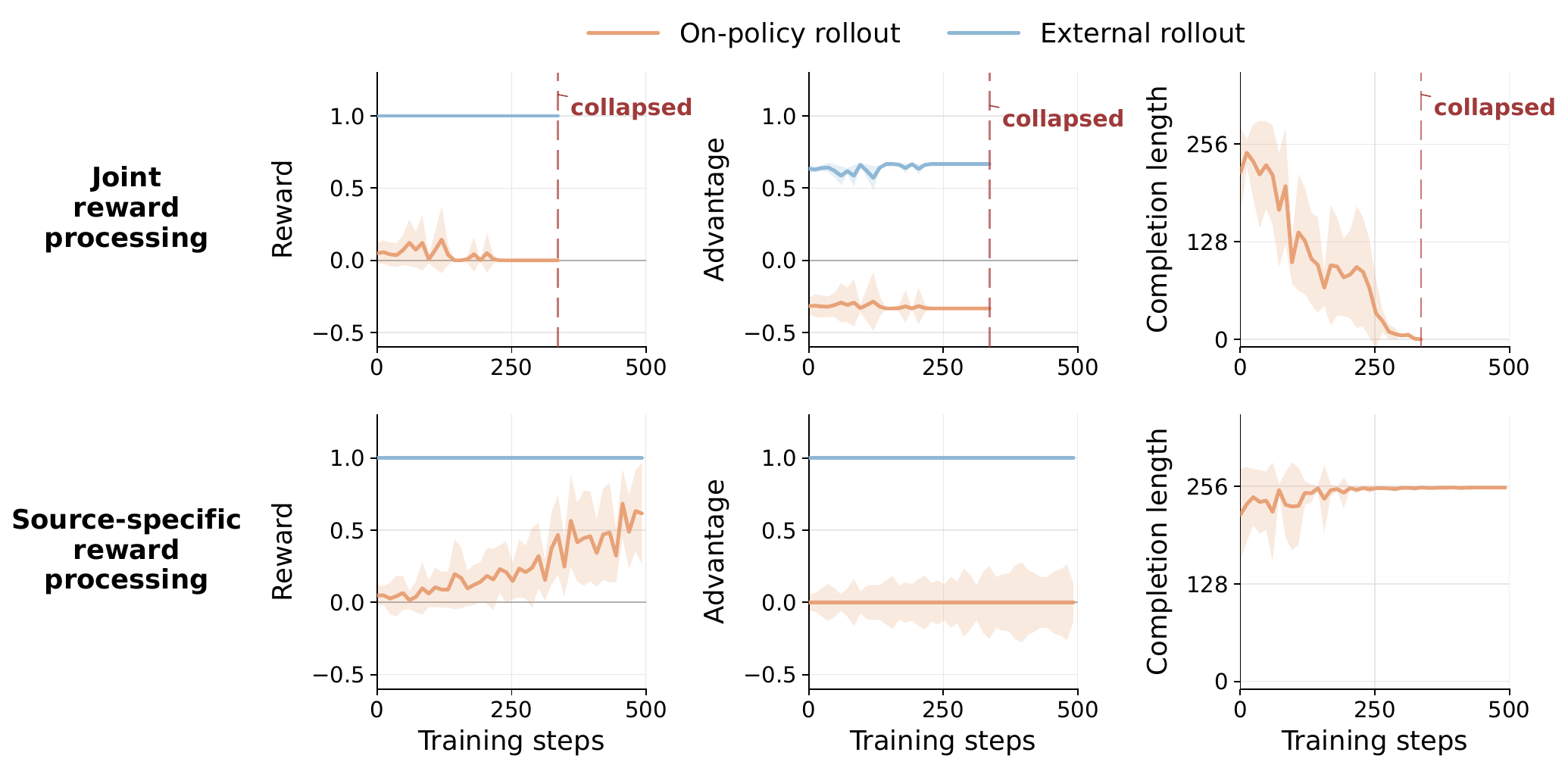}
\caption{
Reward and on-policy training dynamics under joint and source-specific reward processing on Sudoku.
Under joint reward processing (top), the reward gap between external and on-policy rollouts is accompanied by consistently negative on-policy advantages and progressive shortening of on-policy completions before training collapse.
Under source-specific reward processing (bottom), the mean on-policy advantage remains centered near zero and the on-policy completion length remains stable.
Solid lines and shaded regions indicate the mean and standard deviation, respectively.
}
\label{fig:joint_processing_collapse}
\end{figure*}

A key motivation for ERILS is to supplement on-policy rollouts with higher-reward rollouts generated by a stronger external policy.
We therefore examine whether the length-controlled external rollouts indeed receive higher verifier rewards than the initial on-policy rollouts.
As shown in Fig.~\ref{fig:rollout_source_reward}, the external rollouts receive higher average verifier rewards across all three tasks.
Their average rewards are 0.99, 1.00, and 1.70 on Sudoku, Countdown, and MATH500, respectively, compared with 0.08, 0.20, and 1.40 for the initial on-policy rollouts.
These results confirm that the length-controlled external rollouts used by ERILS receive higher average verifier rewards than the target dLLM's initial on-policy rollouts across all three tasks.

\subsection{Analysis of Collapse under Joint Reward Processing}
\label{sec:analysis_joint_processing_collapse}

To better understand the collapse observed under the mixed configuration with joint reward processing, we track the rewards and advantages of the on-policy and external rollouts, as well as the completion lengths of on-policy rollouts during mixed-rollout training on Sudoku.
Every 12 training steps, we analyze the mixed rollout group used for the update, which consists of four on-policy rollouts and two external rollouts.
For each group, we compute the mean and standard deviation separately over the on-policy rollouts and the external rollouts.

Figure~\ref{fig:joint_processing_collapse} reveals distinct on-policy training dynamics under the two reward-processing strategies.
Under joint reward processing, the on-policy rewards remain low while the external rollouts maintain high verifier rewards.
This reward gap places the on-policy rollouts below the mixed-group reward mean, resulting in consistently negative on-policy advantages.
Over the same period, the average on-policy completion length progressively decreases toward zero, at which point training collapses.

Under source-specific reward processing, a similar reward gap is initially present between the external and on-policy rollouts.
However, because the on-policy rewards are centered within the on-policy subgroup, individual on-policy rollouts receive both positive and negative advantages; their mean is zero by construction, as shown in the figure.
The on-policy completion length remains stable rather than collapsing toward zero.
Under this configuration, the on-policy reward also gradually increases over training, while training remains stable.

Overall, these observations suggest that the collapse under joint reward processing is associated with consistently negative on-policy advantages produced by jointly processing the reward gap between external and on-policy rollouts, together with the decreasing on-policy completion length.
These behaviors are not observed under source-specific reward processing, which computes the on-policy baseline independently of the external-rollout rewards.

\section{Conclusion}
\label{sec:conclusion}

We studied how external rollouts generated by a stronger policy can be integrated into reinforcement learning for diffusion language models.
Our results show that effectively incorporating external rollouts requires careful consideration of rollout length and reward processing.
To address these issues, we proposed ERILS, which combines Rollout Length Control with Source-Specific Reward Processing.
Experiments on Sudoku, Countdown, and MATH500 show that ERILS improves dLLM RL under zero-shot evaluation, with the largest gains on Sudoku and additional improvements in multi-sample evaluation on Countdown and MATH500.
Overall, our findings suggest that rollout construction and reward processing are important design dimensions when incorporating external rollouts into dLLM reinforcement learning.
\section*{Acknowledgements}
\label{sec:acknowledge}

This work was supported by Institute of Information \& communications Technology Planning \& Evaluation (IITP) grant funded by the Korea government (MSIT) ([NO.RS-2021-II211343, Artificial Intelligence Graduate School Program (Seoul National University)], [No.RS-2023-00235293, Development of autonomous driving big data processing, management, search, and sharing interface technology to provide autonomous driving data according to the purpose of usage]).

\section*{Declaration of generative AI and AI-assisted technologies in the manuscript preparation process}

During the preparation of this work, the authors used ChatGPT (OpenAI) to assist with language editing. After using this tool, the authors reviewed and edited the content as needed and take full responsibility for the content of the published article.

\printcredits

\bibliographystyle{cas-model2-names}

\bibliography{cas-refs}



\clearpage
\appendix

\makeatletter
\@addtoreset{figure}{section}
\@addtoreset{table}{section}
\@addtoreset{equation}{section}
\makeatother

\renewcommand{\thefigure}{\thesection.\arabic{figure}}
\renewcommand{\thetable}{\thesection.\arabic{table}}
\renewcommand{\theequation}{\thesection.\arabic{equation}}

\section{Diffusion Language Models Background}
\label{sec:appendix_dllm_background}

We first introduce the notation for masked diffusion language models used throughout this paper.
We follow the standard masked-denoising formulation, where the prompt is kept visible and masking is applied only to the completion tokens~\citep{sahoo2024simple,nie2026large}.

Let $p$ denote a prompt and let $o=(o^{(1)},\ldots,o^{(L)})$ denote a clean completion of length $L$, where $L$ may vary across
examples.
During training, a masking ratio $t\sim\mathcal U(0,1)$ is sampled, and a corrupted completion $\widetilde{o}_t\sim q_t(\cdot\mid o)$ is constructed by independently replacing each completion token with $[\mathrm{MASK}]$ with probability $t$.
The corresponding model input is $z_t=[p;\widetilde{o}_t]$, where $[\cdot;\cdot]$ denotes sequence concatenation and the prompt $p$ remains unchanged.
When $t=0$, $z_0=[p;o]$ is the clean prompt--completion sequence.
When $t=1$, $z_1$ consists of the unchanged prompt followed by a fully masked completion.

Let
\begin{equation}
\mathcal M_t = \left\{k\in\{1,\ldots,L\} : \widetilde{o}_t^{(k)}=[\mathrm{MASK}]\right\}
\end{equation}
denote the set of masked positions within the completion.
The model is trained to reconstruct the original token
$o^{(k)}$
at each position
$k\in\mathcal M_t$.

The resulting denoising objective can be derived as an evidence lower bound (ELBO) on the sequence log-likelihood~\citep{sahoo2024simple,shi2024simplified,nie2026large}.
Under the LLaDA-style linear masking schedule, the corresponding ELBO-style denoising score is
\begin{equation}
\mathcal E_\theta(p,o) = \mathbb E_{t,\widetilde{o}_t}\left[
\frac{1}{t} \sum_{k\in\mathcal M_t} \log\pi_\theta(o^{(k)}\mid z_t)
\right],
\label{eq:elbo_objective}
\end{equation}
where $t\sim\mathcal U(0,1)$, $\widetilde{o}_t\sim q_t(\cdot\mid o)$, and $z_t=[p;\widetilde{o}_t]$.
Training minimizes the denoising loss, $-\mathcal{E}_\theta(p,o)$.

At inference time, generation is initialized with a fully masked completion of predefined length $L_{\mathrm{gen}}$.
Let
\begin{equation}
1=\tau_S>\tau_{S-1}>\cdots>\tau_0=0
\end{equation}
denote a discrete schedule of $S$ reverse-denoising steps over the continuous diffusion-time interval.

Generation begins from a fully masked completion:
\begin{equation}
z_{\tau_S} = \left[p; \underbrace{
[\mathrm{MASK}],\ldots,[\mathrm{MASK}]
}_{L_{\mathrm{gen}}}
\right],
\end{equation}
where the prompt remains visible.
At reverse-denoising step $s$, let $z_{\tau_s}=[p;\widetilde{o}_{\tau_s}]$ denote the current partially denoised sequence, and let
\begin{equation}
\mathcal M_s = \left\{
k\in\{1,\ldots,L_{\mathrm{gen}}\} : \widetilde{o}_{\tau_s}^{(k)} = [\mathrm{MASK}]
\right\}
\end{equation}
denote the remaining masked positions within the completion.
Given $z_{\tau_s}$, the dLLM predicts a categorical distribution over the vocabulary at each masked position:
\begin{equation}
\pi_\theta^{(k)}(\cdot\mid z_{\tau_s}), \qquad k\in\mathcal M_s.
\end{equation}

Under the confidence-based decoding strategy used by LLaDA and the dLLM RL baselines~\citep{nie2026large,wang2026spg}, a token is sampled from the predicted distribution at each masked position, and its confidence is defined as the probability assigned to the sampled token by the model.
Tokens with the highest confidence are retained, while the remaining completion positions stay masked for subsequent refinement.
Repeating this procedure yields the reverse-denoising sequence
\begin{equation}
z_{\tau_S} \rightarrow z_{\tau_{S-1}} \rightarrow \cdots \rightarrow z_{\tau_0} = [p;\hat{o}],
\end{equation}
where $\hat{o}$ denotes the generated completion.
\section{External Rollout Construction}
\label{sec:appendix_rollout_construction}

We construct a fixed pool of external rollouts using Qwen3-30B-A3B-Instruct-2507.
The rollout pool is generated before training and kept fixed throughout optimization.
The default ERILS configuration requires two external rollouts per prompt.
For our experiments, we pre-generate six candidate external rollouts per prompt so that the same pool can also support analyses that vary the number of external rollouts.
Each experiment uses only the number of external rollouts specified by its rollout configuration.

The rollout-construction procedure is selected at the task level.
For Countdown and MATH500, we directly generate the final completion using a length-control instruction.
For Sudoku, we use a two-stage rewriting procedure: the first stage generates a detailed solution, and the second stage rewrites it into a compact completion while preserving the final answer.
The verifier reward used for training is always recomputed on the final completion after the applicable length-control procedure.

\subsection{External-Policy Sampling Configuration}
\label{sec:appendix_external_sampling}

External rollouts are generated using Qwen3-30B-A3B-Instruct-2507 as a fixed external policy.
We serve the model with an OpenAI-compatible vLLM backend and query it with the rollout-generation script.
The serving backend is used only for external rollout construction; the external policy is not updated during ERILS training.

Table~\ref{tab:appendix_external_rollout_generation} summarizes the sampling settings used to construct the external-rollout pool.
Unless otherwise specified, the same sampling settings are used across tasks.
For Sudoku, the external policy uses a maximum of 4,096 generation tokens because uncontrolled external rollouts are substantially longer than those of the other tasks.
Countdown and MATH500 use the default limit of 2,048 tokens.

\begin{table}[t]
\centering
\caption{
External-policy sampling settings used to construct the pre-generated external-rollout pool.
The default ERILS configuration uses two external rollouts per prompt; six are generated to support experiments with different rollout configurations.
}
\label{tab:appendix_external_rollout_generation}
\resizebox{\columnwidth}{!}{
\begin{tabular}{ll}
\toprule
Configuration & Value \\
\midrule
External policy & Qwen3-30B-A3B-Instruct-2507 \\
Sampling temperature & 1.0 \\
Completions per prompt & 2 (default ERILS); 6 (pre-generated pool) \\
Maximum generation tokens & 2,048 (Countdown/MATH500); 4,096 (Sudoku) \\
Dataset/order seed & 42 \\
\bottomrule
\end{tabular}
}
\end{table}

For the pre-generated experimental pool, six candidate rollouts are constructed for each prompt.
For Countdown and MATH500, the candidate external rollouts are generated using the length-control instruction.
For the two-stage Sudoku construction, the first stage generates the corresponding detailed completions, and the second stage rewrites each of them into a compact completion.
The default ERILS configuration uses two of these external rollouts per prompt.

\subsection{Completion Format and Length-Control Prompts}
\label{sec:appendix_length_control_prompts}

External rollout construction separates two conceptually different prompt components:

\begin{enumerate}
    \item a task prompt that specifies the problem, task constraints, and required
    answer format; and
    \item a length-control instruction that encourages a shorter completion.
\end{enumerate}

Let \(P_{\mathrm{task}}\) denote the task prompt and \(P_{\mathrm{len}}\) denote the length-control instruction. Countdown and MATH500 use direct generation with the concatenated prompt \(P_{\mathrm{task}} \Vert P_{\mathrm{len}}\).
Sudoku uses \(P_{\mathrm{task}}\) to obtain an initial solution and then applies a separate rewrite prompt containing the initial completion and \(P_{\mathrm{len}}\).

All final external rollouts follow the XML-style completion format:
\begin{center}
\small
\begin{tabular}{@{}l@{}}
\texttt{<reasoning>}~\ldots~\texttt{</reasoning>}\\
\texttt{<answer>}~\ldots~\texttt{</answer>}.
\end{tabular}
\end{center}
For MATH500, the content of the answer field additionally contains the final answer in \verb|\boxed{...}|.

\subsubsection{Task Prompts}

The following prompts contain only the original task instructions and output requirements.
They do not include the length-control instruction.
Braces indicate fields filled by the rollout-generation script.

\begin{qualresponsebox}{Sudoku task prompt}
\small
Please solve the following 4x4 Sudoku puzzle. The puzzle is provided as a 16-character string reading left-to-right, top-to-bottom, where '0' represents empty cells.

Rules:
- Fill empty cells with digits 1-4
- Each row must contain digits 1-4 exactly once
- Each column must contain digits 1-4 exactly once
- Each 2x2 box must contain digits 1-4 exactly once

Important: Your solution must be a COMPLETE 16-character string with only the digits 1-4, representing your final solved grid.

Respond in this exact format:
<reasoning>
Your step-by-step solving process
</reasoning>
<answer>
[16-character solution string with no spaces or separators]
</answer>

Solve the following Sudoku puzzle: \texttt{\{puzzle\}}
\end{qualresponsebox}

\begin{qualresponsebox}{Countdown task prompt}
\small
Respond in the following format:
<reasoning>
...
</reasoning>
<answer>
...
</answer>

Using only the numbers \texttt{\{numbers\}}, create an arithmetic expression that evaluates to exactly \texttt{\{target\}}. You must use all numbers from the list, and each number must be used exactly once. You may use the operations +, -, *, and / as needed. After reasoning, provide only your final expression inside <answer></answer> tags without including an equals sign or the target number. For example, if the numbers are [2, 3, 4] and the target is 5, a valid answer is: <answer>
2*4-3
</answer>
\end{qualresponsebox}

\begin{qualresponsebox}{MATH500 task prompt}
\small
Respond in the following format:
<reasoning>
...
</reasoning>
<answer>
...
</answer>

You are a math expert. You will be given a question to solve. Solve it step by step. Wrap the final answer in a \textbackslash boxed\{\}.

\texttt{\{problem\}}
\end{qualresponsebox}

\subsubsection{Rollout Length Control Prompts}

\paragraph{Direct generation with a length-control instruction.}
The following instruction is appended verbatim to the task prompt used for on-policy rollout generation.
In the main experiments, \texttt{\{target\_total\_tokens\}} is set to 256.
This value serves as a practical reference for completion length and is not used as a truncation boundary.

\begin{qualresponsebox}{Length-control instruction}
\small
[Instruction] Return the full response, including <reasoning> and <answer>, in about \texttt{\{target\_total\_tokens\}} tokens total. Keep <reasoning> compact and focused, ideally no more than 6 short sentences. Do not add repetition, extra narration, or text outside the required tags.
\end{qualresponsebox}

The complete external-policy prompt consists of the original task prompt followed by this instruction.

\paragraph{Rewriting an initially generated completion.}
When direct generation does not reliably produce completions with lengths comparable to $L_{\mathrm{gen}}$, the external policy first generates a completion from the original task prompt.
The original task prompt and the generated completion are then inserted into the following rewrite prompt, followed by the same length-control instruction.

\begin{qualresponsebox}{Rewrite prompt}
\small
\texttt{\{task\_prompt\}}

[Previous detailed solution]
\texttt{\{initial\_completion\}}

[Instruction] Rewrite the solution into a compact final response.
Use the previous detailed solution as the primary reference and preserve the final answer.
Return exactly in the required XML-style format with <reasoning> and <answer> tags.
Keep <reasoning> concise, non-repetitive, and focused only on the decisive steps.
Do not mention that you are rewriting or summarizing a previous answer.
Do not add any text outside the required tags.

\texttt{\{length\_control\_instruction\}}
\end{qualresponsebox}

In both cases, only the final completion is retained as the length-controlled external rollout and added to the fixed external-rollout pool.
For a rewritten completion, the task verifier is applied again to the final completion, and the reward assigned to the initially generated completion is not reused.

\subsection{Filtering and Usability for the Default Configuration}
\label{sec:appendix_rollout_filtering}

\begin{table}[t]
\centering
\caption{
External rollout filtering statistics.
"Default usable" counts prompts with at least two completions of length at most 500 tokens, matching the default ERILS configuration with \(G_{\mathrm{ext}}=2\).
}
\label{tab:appendix_rollout_filtering}
\footnotesize
\setlength{\tabcolsep}{3pt}
\begin{tabular}{@{}lrrrrr@{}}
\toprule
Task
& \makecell{Source\\prompts}
& \makecell{Generated\\completions}
& \makecell{Completions\\\(>500\)}
& \makecell{Default\\usable}
& \makecell{All-six\\valid} \\
\midrule
Sudoku    & 10,000 & 60,000 & 269   & 10,000 & 9,794 \\
Countdown & 10,000 & 60,000 & 3,330 & 9,764  & 8,742 \\
MATH500   & 7,500  & 45,000 & 4,023 & 7,177  & 6,126 \\
\bottomrule
\end{tabular}
\end{table}

All external completion lengths are measured using the Qwen3 tokenizer over the generated completion only, excluding the task prompt.
The same definition is used for the completion-length statistics and the 500-token filtering threshold.
After generation, external rollouts are filtered and reordered solely according to completion length; verifier rewards and answer correctness are not used for either operation.
Rollouts longer than 500 tokens are treated as unusable for the length-controlled external rollout setting.
Because the default ERILS configuration uses $G_{\mathrm{ext}}=2$, a prompt is considered usable when it has at least two external rollouts of at most 500 tokens.
For such prompts, usable rollouts are moved to the front of the rollout list, and the first two are used as the external rollouts in the default configuration.

Because we pre-generate six external rollouts per prompt for the broader experimental pool, we additionally report an all-six-valid criterion for diagnostic purposes.
This criterion is stricter than what the default ERILS configuration requires.

For Sudoku and Countdown, the external-rollout pool contains 10,000 prompts.
This exceeds the number of distinct prompt groups used during the main ERILS runs, so training does not cycle through the external prompt pool.
Any repeated use of a selected prompt group results from the multiple optimizer updates applied to each rollout batch rather than an insufficient pool size.
For MATH500, 4,023 of the 45,000 generated rollouts exceed 500 tokens; nevertheless, 7,177 of the 7,500 prompts satisfy the default usability criterion of having at least two usable external rollouts.

\subsection{Sudoku Two-Stage Rewriting Analysis}
\label{sec:appendix_sudoku_rewriting_analysis}

We examine whether the Sudoku two-stage rewriting procedure degrades the correctness of the original external rollouts.
The analysis covers 60,000 paired first- and second-stage completions from the full external-rollout pool.

\begin{table}[t]
\centering
\caption{
Sudoku external-rollout statistics before and after two-stage rewriting.
A correct completion receives a verifier reward of 1.
}
\label{tab:appendix_sudoku_rewriting}
\resizebox{\columnwidth}{!}{
\begin{tabular}{lrrr}
\toprule
 & Mean completion length & Mean verifier reward & Correct completions \\
\midrule
First stage  & 2,432.83 & 0.9724 & 58,219 (97.03\%) \\
Second stage & 229.60   & 0.9903 & 59,042 (98.40\%) \\
\bottomrule
\end{tabular}
}
\end{table}

As shown in Table~\ref{tab:appendix_sudoku_rewriting}, all 58,219 first-stage correct completions remain correct according to the verifier after rewriting, while 823 additional completions become correct.
Because the verifier evaluates only the cells that were originally empty, we additionally compare the generated boards with the complete reference solutions.
Under this stricter comparison, only two of the 58,219 first-stage correct completions become incorrect after rewriting; 99.9966\% remain correct.
Neither case appears among the external rollouts used for default ERILS training.
These results show that the two-stage rewriting procedure does not meaningfully degrade the correctness of the external rollouts.

\subsection{Prompt Order Alignment}
\label{sec:appendix_prompt_order_alignment}

In our implementation, external rollouts are indexed by their training prompt and ordered according to the prompt exposure order used by the training code.
The external rollouts used by ERILS therefore correspond to the prompts used for on-policy rollout generation at each training step.
We use the same training-prompt order for the on-policy baseline and ERILS, preventing differences in prompt order from affecting the comparison.
\section{Experimental Details}
\label{sec:appendix_experimental_details}

This section provides the training and evaluation details for ERILS and for the baselines that we reproduce.
Additional details on the baseline results are provided in Section~\ref{sec:appendix_baselines}.
Our implementation follows the general experimental protocol of prior dLLM reinforcement-learning studies, particularly SPG~\citep{wang2026spg}, while using the rollout configuration described in Section~\ref{sec:method}.

\subsection{Datasets and Data Splits}
\label{sec:appendix_datasets}

We evaluate ERILS on Sudoku~\citep{arel_sudoku_2025}, Countdown~\citep{tinyzero}, and MATH500~\citep{lightman2024let}.
Table~\ref{tab:appendix_dataset_statistics} summarizes the training and evaluation splits used for each task.

\begin{table}[t]
\centering
\caption{Dataset statistics. We report the number of original prompts available for training and the number of prompts used for final evaluation. No separate validation split was used.}
\label{tab:appendix_dataset_statistics}
\begin{tabular}{lcc}
\toprule
Task & Training & Evaluation \\
\midrule
Sudoku    & 1,000,000 & 256 \\
Countdown & 240,632 & 256 \\
MATH500   & 7,500 & 500 \\
\bottomrule
\end{tabular}
\end{table}

We adopt the dataset preprocessing, task-specific reward functions, and verifier implementations from the released SPG codebase~\citep{wang2026spg}.
For Countdown and MATH500, these implementations follow the experimental protocols introduced in d1 and subsequently used by wd1~\citep{zhao2026d1,tang2026wd1}.
For Sudoku, we adopt the solution-disjoint data split constructed by SPG.
We retain the released task-specific preprocessing and verification logic without modification, while using zero-shot prompts for all three tasks.

\paragraph{Sudoku.}
We adopt the solution-disjoint split and task-specific verifier released with SPG~\citep{wang2026spg}.
SPG constructs this split from the Sudoku dataset used by d1 by assigning complete solution grids to disjoint training and evaluation sets.
We use the resulting split and verification code without modification.
Our prompting protocol differs from SPG: whereas SPG uses three solved demonstrations, we train and evaluate all models in the zero-shot setting.

\paragraph{Countdown.}
We use the dataset construction, preprocessing, and verifier distributed with SPG, which follow the three-number Countdown protocol of d1~\citep{zhao2026d1,wang2026spg}.
Training uses the corresponding synthetic training set, and evaluation uses the 256 synthetic three-number problems introduced by d1.
We do not modify the task-specific reward or verification logic.

\paragraph{MATH500.}
We use the mathematical-reasoning preprocessing, answer parser, and reward implementation distributed with SPG, following the MATH protocol of d1~\citep{zhao2026d1,wang2026spg}.
Training uses the MATH training split, and evaluation is conducted on MATH500~\citep{lightman2024let}.
We retain the released task-specific reward and answer-verification logic without modification.

\subsection{Reward Functions and Answer Verification}
\label{sec:appendix_rewards}

We directly use the task-specific reward functions and verifier implementations released with SPG~\citep{wang2026spg}.
These implementations follow d1 and wd1 for Countdown and MATH500 and use the Sudoku verifier distributed with the SPG codebase.
The same verifier code is applied to on-policy rollouts, external rollouts, and evaluation generations.
We do not introduce task-specific modifications to the reward or answer-parsing logic.

\paragraph{Sudoku.}
The verifier parses the generated $4\times4$ grid and computes the fraction of originally empty cells filled with the correct values.
Pre-filled cells are excluded from the reward calculation so that the reward measures puzzle-solving performance rather than copying of the input grid.
Malformed outputs and missing cells are treated as incorrect at their corresponding positions.
For multi-sample evaluation, we report best-of-$k$ completion accuracy using this cell-level score.
For deterministic single-completion evaluation, we report the completion accuracy of the single generated completion.

\paragraph{Countdown.}
The verifier extracts an arithmetic expression from the completion and checks both its numerical value and its use of the provided numbers.
A reward of $1.0$ is assigned when the expression reaches the target using exactly the available input numbers.
A reward of $0.1$ is assigned when the expression uses the correct multiset of input numbers but does not reach the target, and zero is assigned otherwise.
Malformed expressions, expressions containing unavailable numbers, and expressions that reuse a number more often than permitted receive zero reward.
For evaluation, a completion is counted as correct only when it reaches the target using exactly the provided numbers.

\paragraph{MATH500.}
The verifier parses the completion according to the prescribed \texttt{<answer>} and \verb|\boxed{...}| format.
Following the adopted implementation, the training reward combines format and answer-correctness components.
The format reward is $1.00$ when a boxed answer appears inside the \texttt{<answer>} field, $0.75$ when the \texttt{<answer>} field is present without a boxed expression, $0.50$ when a boxed expression appears without the \texttt{<answer>} field, and $0.25$ otherwise.
An additional correctness reward of $2.0$ is assigned when the extracted boxed answer matches the reference answer.
For evaluation, a completion is counted as correct only when the extracted final answer matches the reference answer; the format component does not contribute to the reported accuracy.

For length-controlled external rollouts, including completions produced by the optional second-stage Sudoku rewrite, the verifier is applied again to the final completion.
Rewards from the uncontrolled or first-stage completion are not reused.

\subsection{Models and Parameter-Efficient Fine-Tuning}
\label{sec:appendix_models}

We use LLaDA-8B-Instruct~\citep{nie2026large} as the target dLLM and Qwen3-30B-A3B-Instruct-2507~\citep{qwen3technicalreport} as the fixed external policy.
The external policy is used only to generate candidate completions and is not updated during ERILS training.
The target dLLM is fine-tuned with Low-Rank Adaptation (LoRA), applied to the attention and MLP projection layers.
Table~\ref{tab:appendix_model_hyperparameters} summarizes the model-loading and LoRA configuration.

\begin{table}[t]
\centering
\caption{Model-loading and LoRA fine-tuning configuration.}
\label{tab:appendix_model_hyperparameters}
\begin{tabular}{ll}
\toprule
Configuration & Value \\
\midrule
Target dLLM & LLaDA-8B-Instruct \\
External policy & Qwen3-30B-A3B-Instruct-2507 \\
LoRA rank $r$ & 128 \\
LoRA scaling $\alpha$ & 64 \\
LoRA dropout & 0.05 \\
LoRA target modules & Attention and MLP projections \\
Target dLLM loading & 4-bit NF4, bfloat16 compute \\
\bottomrule
\end{tabular}
\end{table}

\subsection{ERILS Training Configuration}
\label{sec:appendix_training_configuration}

This section summarizes the training configuration used for ERILS.
Unless otherwise specified, each rollout group contains six rollouts.
The default configuration uses four on-policy rollouts and two length-controlled external rollouts for each prompt.
Source-specific reward processing follows Section~\ref{sec:mixed_learning}, and the resulting on-policy and external-rollout contributions are combined over the mixed rollout group as in Eq.~\ref{eq:erils_update}.

The diffusion-compatible surrogate objective follows the SPG formulation for both rollout sources.
For on-policy rollouts, we use the same confidence-based block masking strategy as the corresponding SPG baseline.
Because external rollouts are not generated by the target dLLM, the rollout-time confidence information used by this masking strategy is unavailable; we therefore sample masked positions uniformly at random for external rollouts.
On-policy rollouts are generated from the current target dLLM using confidence-based decoding with low-confidence remasking and a training generation length of 256 tokens.
The remaining optimization and rollout-generation settings are reported in Table~\ref{tab:appendix_training_hyperparameters}.

For the main benchmark experiments, Sudoku and Countdown are trained for 2,500 steps, while MATH500 is trained for 6,000 steps.
Analysis experiments on Sudoku are trained for 4,000 training steps unless otherwise specified.

\begin{table}[t]
\centering
\caption{ERILS training hyperparameters.}
\label{tab:appendix_training_hyperparameters}
\resizebox{\columnwidth}{!}{
\begin{tabular}{ll}
\toprule
Hyperparameter & Value \\
\midrule
Optimizer & AdamW \\
Learning rate & $3 \times 10^{-6}$ \\
Adam $\beta_1$ & 0.9 \\
Adam $\beta_2$ & 0.99 \\
Weight decay & 0.1 \\
Learning-rate schedule & Constant with warmup \\
Maximum gradient norm & 0.2 \\
Per-device prompt-group batch size & 6 \\
Gradient accumulation steps & 1 \\
GPU model & NVIDIA GeForce RTX 3090 \\
Training GPUs & Main experiments: 8; Analysis: 4 \\
Rollouts per prompt & 6 \\
On-policy rollouts per group & 4 \\
External rollouts per group & 2 \\
Training generation length & 256 \\
Maximum prompt length & 200 \\
Diffusion steps for on-policy rollout generation & 128 \\
Block length & 32 \\
On-policy rollout temperature & 1.0 \\
Remasking strategy & Low confidence \\
Optimizer updates per rollout batch & 12 \\
\bottomrule
\end{tabular}
}
\end{table}

\subsection{Evaluation Protocol}
\label{sec:appendix_evaluation_protocol}

All evaluations use zero-shot prompts.
We consider two generation settings: deterministic single-completion evaluation and multi-sample evaluation.

For deterministic single-completion evaluation, we use temperature 0.0 and generation lengths $L_{\mathrm{gen}}\in\{128,256,512\}$.
The number of diffusion steps is set to $L_{\mathrm{gen}}/2$, corresponding to 64, 128, and 256 steps, respectively.
One completion is generated for each prompt.

For multi-sample evaluation, we follow the sampling protocol used in SPG~\citep{wang2026spg}.
We use temperature 0.9 and generate four completions for each prompt.
For Countdown and MATH500, we report Pass@$k$ for $k\in\{1,2,3,4\}$.
For Sudoku, we report best-of-$k$ completion accuracy, where completion accuracy is the fraction of originally empty grid positions filled with the correct digits.
For each prompt, the four completions are generated using fixed sampling seeds 1001, 1002, 1003, and 1004.
For each value of $k$, Pass@$k$ and best-of-$k$ are computed using the completions generated with the first $k$ seeds in this fixed order.
Unless otherwise specified, multi-sample evaluation uses $L_{\mathrm{gen}}=256$.

For the main benchmark, checkpoints are evaluated every 100 training steps, beginning at step 1,500 for Sudoku and Countdown and at step 2,000 for MATH500.
Following the checkpoint reporting practice used in prior dLLM RL studies~\citep{zhao2026d1,wang2026spg}, we report the best observed checkpoint result separately for each task and reported evaluation configuration.
No separate validation split is used, and different reported evaluation configurations may therefore correspond to different checkpoints.

For analysis experiments, checkpoints are evaluated every 100 training steps on a fixed 128-example subset of the Sudoku evaluation set to track performance over training.

\begin{table}[t]
\centering
\caption{Evaluation-generation settings for multi-sample and deterministic single-completion evaluation.}
\label{tab:appendix_evaluation_hyperparameters}
\resizebox{\columnwidth}{!}{
\begin{tabular}{lll}
\toprule
Configuration & Multi-sample & Deterministic single-completion \\
\midrule
Prompting & Zero-shot & Zero-shot \\
Generation length & 256 & 128, 256, 512 \\
Sampling temperature & 0.9 & 0.0 \\
Samples per prompt & 4 & 1 \\
Reported metric & Pass@$k$ / best-of-$k$ & Accuracy \\
Decoding strategy & Confidence-based & Confidence-based \\
Remasking strategy & Low confidence & Low confidence \\
Diffusion steps & 128 & Generation length / 2 \\
Block length & 32 & 32 \\
\bottomrule
\end{tabular}
}
\end{table}

\subsection{Baseline Results}
\label{sec:appendix_baselines}

For multi-sample evaluation in Table~\ref{tab:multi_sample_evaluation}, the Countdown and MATH500 baseline results are taken from prior reported results under the corresponding sampling protocol.
All Sudoku results are obtained from our zero-shot evaluations under the same four-sample protocol used for ERILS.
All ERILS results are obtained from our experiments.

For deterministic single-completion evaluation in Table~\ref{tab:main_benchmark}, baseline results are taken from the corresponding original papers unless otherwise specified.
The d1 and wd1 values are taken from their original papers.
The LLaDA-8B-Instruct baseline is taken from d1, while the LLaDA~1.5, UniGRPO, and SPG results for Countdown and MATH500 are taken from SPG~\citep{wang2026spg}.
Because the original SPG paper evaluates Sudoku using three solved demonstrations, we reproduce SPG under the same zero-shot protocol used for ERILS.
The $\dagger$ symbol in Table~\ref{tab:main_benchmark} denotes this zero-shot SPG reproduction.

\subsection{Protocol for Rollout Length and Reward Analysis}
\label{sec:appendix_rollout_diagnostic}

The analyses of external rollout length and verifier reward reported in Figs.~\ref{fig:external_rollout_length} and~\ref{fig:rollout_source_reward} use the following diagnostic protocol.
We examine whether Rollout Length Control reduces external completion length and whether the resulting rollouts retain higher verifier rewards than on-policy rollouts generated by the initial target dLLM.

To ensure a consistent comparison across tasks with different training-set sizes, we randomly select 1,000 training prompts from each task.
Within each task, the same prompt subset is used for all three conditions: uncontrolled external rollouts, length-controlled external rollouts, and initial on-policy rollouts.
For each prompt and condition, we evaluate six rollouts, resulting in 6,000 rollouts per condition for each task.

\subsection{Supervised Fine-Tuning Baseline}
\label{sec:appendix_sft}

For the supervised fine-tuning baseline, we use the masked-denoising fine-tuning recipe released by d1~\citep{zhao2026d1}.
The SFT dataset consists only of length-controlled external rollouts from the pre-generated external-rollout pool described in Appendix~\ref{sec:appendix_rollout_construction}.
Unlike ERILS, which uses four on-policy rollouts and two external rollouts per prompt, SFT uses six external rollouts per prompt and does not generate on-policy rollouts during training.
The target model, LoRA configuration, precision setting, training split, and hardware setup match ERILS.
We evaluate SFT on the same fixed 128-example Sudoku evaluation subset used in the analysis experiments.

\section{Sensitivity to Checkpoint Selection}
\label{sec:appendix_checkpoint_sensitivity}

The main benchmark results report the best observed checkpoint separately for each task and reported evaluation configuration, following the checkpoint reporting practice described in Section~\ref{sec:exp_setup}.
To examine whether the reported performance depends strongly on this checkpoint selection, we additionally evaluate ERILS at the final training checkpoint.
For Sudoku and Countdown, this corresponds to step 2,500, and for MATH500, to step 6,000.
Because MATH500 is trained for substantially more steps, we also report the results at step 4,000.

\begin{table}[t]
\centering
\caption{
Multi-sample evaluation of fixed ERILS checkpoints.
Sudoku reports best-of-$k$ completion accuracy, while Countdown and MATH500 report Pass@$k$.
}
\label{tab:appendix_last_checkpoint_multi}
\resizebox{\columnwidth}{!}{
\begin{tabular}{lrrrrr}
\toprule
Task & Step & $k=1$ & $k=2$ & $k=3$ & $k=4$ \\
\midrule
Sudoku    & 2,500 & 90.5 & 94.6 & 96.3 & 97.7 \\
Countdown & 2,500 & 62.1 & 75.8 & 79.3 & 83.2 \\
MATH500   & 4,000 & 33.4 & 45.0 & 52.0 & 56.0 \\
MATH500   & 6,000 & 34.2 & 41.8 & 47.2 & 51.6 \\
\bottomrule
\end{tabular}
}
\end{table}

\begin{table}[t]
\centering
\caption{
Deterministic single-completion evaluation of fixed ERILS checkpoints across generation lengths.
}
\label{tab:appendix_last_checkpoint_single}
\resizebox{\columnwidth}{!}{
\begin{tabular}{lrrrr}
\toprule
Task & Step & 128 & 256 & 512 \\
\midrule
Sudoku    & 2,500 & 90.67 & 90.23 & 87.89 \\
Countdown & 2,500 & 61.33 & 67.58 & 67.58 \\
MATH500   & 4,000 & 33.00 & 36.60 & 34.60 \\
MATH500   & 6,000 & 31.40 & 33.00 & 37.20 \\
\bottomrule
\end{tabular}
}
\end{table}

The fixed-checkpoint results show that the multi-sample evaluation retains strong ERILS performance without selecting a separate checkpoint for each evaluation configuration.
The deterministic single-completion results exhibit larger variation across checkpoints, particularly on MATH500.
Overall, multi-sample evaluation is more robust to checkpoint choice in these experiments, while deterministic single-completion evaluation is more sensitive to the particular checkpoint used for reporting.

\section{Cost Analysis}
\label{sec:appendix_cost_analysis}

ERILS uses an external policy to construct external rollouts before training, introducing an additional rollout-generation cost compared with the on-policy SPG baseline.
Because external-rollout construction is performed before training, the external policy does not need to be loaded concurrently with the target dLLM during ERILS training.
We therefore compare the wall-clock cost of ERILS and SPG under the main training configuration.
External-rollout generation and the main training runs both use eight NVIDIA GeForce RTX 3090 GPUs.
Table~\ref{tab:appendix_time_cost} reports external-rollout generation time, training time, and total wall time for each task.

\begin{table}[t]
\centering
\caption{
Wall-clock cost of SPG and ERILS.
External-rollout generation is performed only for ERILS.
The external-rollout generation times correspond to the two external rollouts per prompt used in the default ERILS configuration.
}
\label{tab:appendix_time_cost}
\resizebox{\columnwidth}{!}{
\begin{tabular}{llrr}
\toprule
Task & Cost & SPG & ERILS \\
\midrule
Sudoku
& External-rollout generation & -- & 4h 38m 53s \\
& Training & 10h 33m 28s & 8h 17m 52s \\
& \textbf{Total wall time} & \textbf{10h 33m 28s} & \textbf{12h 56m 45s} \\
\midrule
Countdown
& External-rollout generation & -- & 18m 58s \\
& Training & 9h 57m 21s & 7h 55m 10s \\
& \textbf{Total wall time} & \textbf{9h 57m 21s} & \textbf{8h 14m 08s} \\
\midrule
MATH500
& External-rollout generation & -- & 1h 01m 40s \\
& Training & 24h 19m 21s & 19h 55m 23s \\
& \textbf{Total wall time} & \textbf{24h 19m 21s} & \textbf{20h 57m 03s} \\
\bottomrule
\end{tabular}
}
\end{table}

As shown in Table~\ref{tab:appendix_time_cost}, ERILS requires less training time than SPG on all three tasks.
This reduction partly reflects the rollout composition of ERILS: two of the six rollouts per prompt are external rollouts constructed before training, reducing the number of on-policy rollouts that must be generated during reinforcement learning.
After including external-rollout generation, ERILS has lower total wall time than SPG on Countdown and MATH500.
On Sudoku, the two-stage rewriting procedure increases the cost of external-rollout construction, resulting in additional total wall time, although ERILS still requires less reinforcement-learning training time than SPG.
Overall, using external rollouts does not necessarily increase wall-clock cost: ERILS is faster overall on Countdown and MATH500, while on Sudoku the additional external-rollout generation cost results in a modest increase in total wall time.

\section{Completion-Length Dynamics}
\label{sec:appendix_completion_length_dynamics}

\begin{figure*}[t]
\centering
\includegraphics[width=\textwidth]{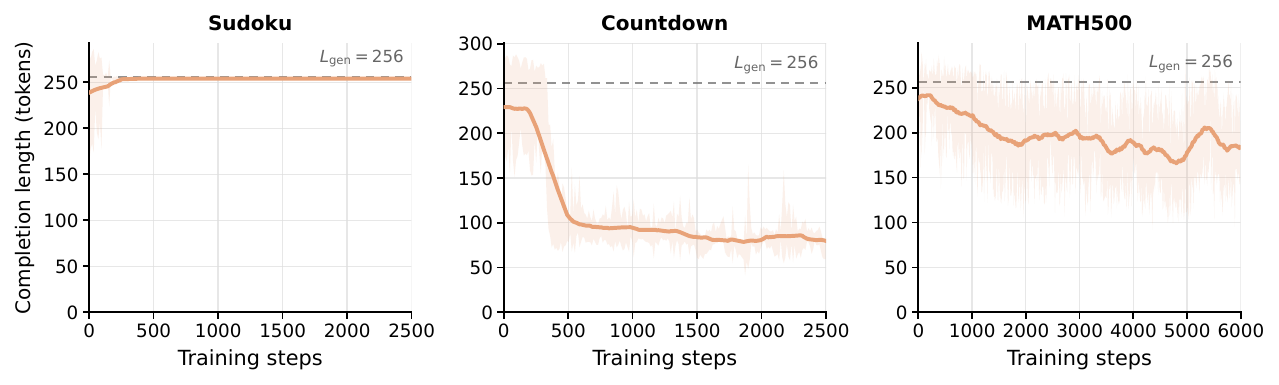}
\caption{
Completion lengths of on-policy rollouts during ERILS training on Sudoku, Countdown, and MATH500.
Solid lines show mean values smoothed using a centered moving average over 25 logged points, with statistics recorded every 12 training steps.
Shaded regions indicate the standard deviation computed from the unsmoothed statistics.
The dashed horizontal line indicates the training generation length of 256 tokens.
}
\label{fig:erils_completion_length_dynamics}
\end{figure*}

Figure~\ref{fig:erils_completion_length_dynamics} shows the completion lengths of the on-policy rollouts during ERILS training.
The completion-length dynamics differ across tasks.
On Sudoku, the on-policy completion length remains close to the training generation length throughout training.
On Countdown, it decreases substantially during the early stage of training and remains shorter thereafter.
MATH500 shows a more gradual decrease with larger variation over training.
Together with the verifier-reward dynamics reported in Section~\ref{sec:reward_dynamics}, these results show that on-policy
verifier rewards increase across tasks despite different completion-length dynamics.

\section{External Rollout Examples}
\label{sec:appendix_external_examples}

We provide representative external rollouts before and after length control.
Each example includes the task input, generated completion, completion length, and verifier reward.

\subsection{Sudoku}

\rollouttitle{Uncontrolled external rollout.}
\begin{lstlisting}[style=rolloutstyle]
Problem:
0432030142033000

Completion:
<reasoning>
We are given a 4x4 Sudoku puzzle as a 16-character string:

"0432030142033000"

We interpret this as a grid row by row:

Row 1: 0 4 3 2
Row 2: 0 3 0 1
Row 3: 4 2 0 3
Row 4: 3 0 0 0

So the grid is:

. 4 3 2
. 3 . 1
4 2 . 3
3 . . .

We need to fill in the blanks with digits 1-4 such that each row, column, and 2x2 box contains all digits 1-4 exactly once.

Step 1: Row 1 is . 4 3 2, so the missing digit is 1. Therefore, cell (1,1) = 1.

Step 2: Row 2 is . 3 . 1, so the missing digits are 2 and 4. Column 1 now has 1, blank, 4, 3, so the missing digit is 2. Therefore, cell (2,1) = 2, and cell (2,3) = 4.

The grid is now:

1 4 3 2
2 3 4 1
4 2 . 3
3 . . .

Step 3: Row 3 is 4 2 . 3, so the missing digit is 1. Therefore, cell (3,3) = 1.

Step 4: Column 3 is 3, 4, 1, blank, so the missing digit is 2. Therefore, cell (4,3) = 2.

Step 5: Row 4 is 3 . 2 ., so the missing digits are 1 and 4. Column 2 is 4, 3, 2, blank, so the missing digit is 1. Therefore, cell (4,2) = 1, and cell (4,4) = 4.

Now the final grid is:

1 4 3 2
2 3 4 1
4 2 1 3
3 1 2 4

Check rows, columns, and 2x2 boxes: each contains digits 1-4 exactly once.

As a 16-character string, the solved grid is 1432234142133124.
</reasoning>
<answer>
1432234142133124
</answer>

Completion length:
1191 tokens

Verifier reward:
1.0
\end{lstlisting}

\rollouttitle{Completion generated with the length control instruction.}
\begin{lstlisting}[style=rolloutstyle]
Problem:
0432030142033000

Completion:
<reasoning>
The puzzle is a 4x4 grid with given clues. Analyze row 1: "0432" has 4,3,2, so missing 1 -> first cell is 1. Row 2: "0301" has 3,1, so missing 1,2,4; column 1 has 1,0,4,0 -> possible values constrained. Column 1: values 1 (from row1), 0, 4, 0 -> missing 2,3. Box 1 (top-left 2x2) has 1,4,3,2 -> complete. Box 2 (top-right): cells (1,3)=3, (1,4)=2, (2,3)=0, (2,4)=1 -> missing 4 -> (2,3)=4. Row 2 now: 0,3,4,1 -> missing 2 -> (2,1)=2. Column 1: 1,2,4,0 -> missing 3 -> (4,1)=3. Row 4: 3,0,0,0 -> missing 1,2,4. Column 2: 4,3,0,0 -> missing 1,2. Box 3 (bottom-left): (3,1)=4, (3,2)=0, (4,1)=3, (4,2)=0 -> missing 1,2. Row 3: 4,2,3,3 -> wait, inconsistency? Wait, row 3: "4203" -> (3,1)=4, (3,2)=2, (3,3)=0, (3,4)=3 -> missing 1 -> (3,3)=1. Then (3,2)=2, (3,3)=1 -> so column 2: (1,2)=4, (2,2)=3, (3,2)=2, (4,2)=? -> missing 1 -> (4,2)=1. Row 4: 3,1,?,? -> missing 2,4. Column 3: (1,3)=3, (2,3)=4, (3,3)=1, (4,3)=? -> missing 2 -> (4,3)=2. Then (4,4)=4. Final row 4: 3,1,2,4. Check column 4: 2,1,3,4 -> valid. All rows, columns, boxes complete.
</reasoning>
<answer>
1432234142133124
</answer>

Completion length:
582 tokens

Verifier reward:
1.0
\end{lstlisting}

\rollouttitle{Length-controlled external rollout after two-stage rewriting.}
\begin{lstlisting}[style=rolloutstyle]
Problem:
0432030142033000

Completion:
<reasoning>
Parse the 16-character string into a 4x4 grid. Row 1 has 0,4,3,2 -- missing 1, so (1,1) = 1. Row 2: 0,3,0,1 -- after (1,1)=1, Box 1 missing 2, so (2,1)=2. Then Row 2 missing 4, so (2,3)=4. Row 3: 4,2,0,3 -- missing 1, so (3,3)=1. Column 2 has 4,3,2, so (4,2)=1. Row 4: 3,1,0,0 -- missing 2,4; Column 3 has 3,4,1, so (4,3)=2; then (4,4)=4. All rows, columns, and boxes now contain 1-4 exactly once.
</reasoning>
<answer>
1432234142133124
</answer>

Completion length:
235 tokens

Verifier reward:
1.0
\end{lstlisting}

\subsection{Countdown}

\rollouttitle{Uncontrolled external rollout.}
\begin{lstlisting}[style=rolloutstyle]
Problem:
Using only the numbers [63, 42, 3], create an arithmetic expression that evaluates to exactly 49.

Completion:
<reasoning>
We are given the numbers [63, 42, 3] and need to create an expression using each number exactly once, with operations +, -, *, and /, to evaluate to exactly 49.

Let's explore possible combinations.

First, note that 63 and 42 are both multiples of 21:
- 63 = 21 * 3
- 42 = 21 * 2

And we have a 3, so perhaps we can use that.

Try: 63 - 42 = 21
Then 21 * 3 = 63, too big.
But 42 / 3 = 14.
Then 63 - 14 = 49. That's promising.

Let's check:
63 - (42 / 3) = 63 - 14 = 49, correct.

Used all numbers: 63, 42, 3, each once.
Used only allowed operations: subtraction and division.

So the expression is: 63 - (42 / 3)

This evaluates to 49.
</reasoning>
<answer>
63 - (42 / 3)
</answer>

Completion length:
297 tokens

Verifier reward:
1.0
\end{lstlisting}

\rollouttitle{Length-controlled external rollout.}
\begin{lstlisting}[style=rolloutstyle]
Problem:
Using only the numbers [63, 42, 3], create an arithmetic expression that evaluates to exactly 49.

Completion:
<reasoning>
We need to use 63, 42, and 3 exactly once with basic operations to get 49.
Try subtracting 42 from 63: 63 - 42 = 21. Then 21 * 3 = 63, too big.
Try 63 - (42 / 3) = 63 - 14 = 49. That works.
Check: 42 / 3 = 14, 63 - 14 = 49. All numbers used once.
</reasoning>
<answer>
63 - 42 / 3
</answer>

Completion length:
149 tokens

Verifier reward:
1.0
\end{lstlisting}

\subsection{MATH500}

\rollouttitle{Uncontrolled external rollout.}
\begin{lstlisting}[style=rolloutstyle]
Problem:
Find the smallest, positive four-digit multiple of $15.$

Completion:
<reasoning>
We are asked to find the smallest positive four-digit multiple of 15.

Step 1: Understand the constraints
- A four-digit number ranges from 1000 to 9999.
- We want the smallest such number that is a multiple of 15.

Step 2: Use divisibility rules for 15
A number is divisible by 15 if and only if it is divisible by both 3 and 5.

- Divisibility by 5: The number must end in 0 or 5.
- Divisibility by 3: The sum of the digits must be divisible by 3.

But instead of checking digit sums, we can use a more direct method: find the smallest four-digit number divisible by 15.

Step 3: Find the smallest four-digit number >= 1000 that is divisible by 15.

We divide 1000 by 15:

1000 / 15 = 66.666...

So the smallest integer greater than or equal to 66.666... is 67.

Now multiply 15 by 67:

15 * 67 = 1005

Step 4: Verify
- 1005 is a four-digit number.
- Ends in 5, so it is divisible by 5.
- Sum of digits: 1 + 0 + 0 + 5 = 6, which is divisible by 3.
- So, it is divisible by 15.

The previous multiple is 15 * 66 = 990, which is a three-digit number, so it is too small.

Thus, 1005 is the smallest four-digit multiple of 15.
</reasoning>
<answer>
\boxed{1005}
</answer>

Completion length:
473 tokens

Verifier reward:
2.75
\end{lstlisting}

\rollouttitle{Length-controlled external rollout.}
\begin{lstlisting}[style=rolloutstyle]
Problem:
Find the smallest, positive four-digit multiple of $15.$

Completion:
<reasoning>
The smallest four-digit number is 1000.
A number divisible by 15 must be divisible by both 3 and 5.
Numbers divisible by 5 end in 0 or 5.
Check if 1000 is divisible by 3: sum of digits is 1 + 0 + 0 + 0 = 1, not divisible by 3.
Try the next multiple of 5 after 1000: 1005.
Sum of digits of 1005 is 1 + 0 + 0 + 5 = 6, divisible by 3.
Thus, 1005 is divisible by both 3 and 5, so divisible by 15.
</reasoning>
<answer>
\boxed{1005}
</answer>

Completion length:
194 tokens

Verifier reward:
2.75
\end{lstlisting}
\section{Qualitative Generation Examples}
\label{sec:appendix_generation_examples}

We compare completions generated by SPG-trained and ERILS-trained models on the same questions.
All examples use a generation length of 256 tokens and 128 diffusion steps.
Green text marks correct computations or outputs accepted by the verifier, while red text marks errors that lead to incorrect completions.

\subsection{Representative Improvements}
The following examples compare cases in which SPG fails while ERILS succeeds on the same question.
On Sudoku, the example highlights complete candidate construction.
On Countdown, the examples illustrate input accounting and operation composition.
On MATH500, the examples show reasoning--answer consistency and algebraic consistency.

\subsubsection{Sudoku: Complete Candidate Construction}
\begin{figure}[pos=t]
\centering
\begin{minipage}{0.98\columnwidth}
\begin{qualresponsebox}{Question}
\small
\textbf{Question:} Solve the following Sudoku puzzle:
\texttt{0013010024011020}

\medskip
\textbf{Reference answer:} \texttt{4213314224311324}
\end{qualresponsebox}
\begin{qualresponsebox}{SPG}
\small
\textbf{Reasoning.}
To solve this 4x4 Sudoku puzzle, we need to fill in the empty cells with digits
1--4 such that each row, column, and $2\times2$ box contains all digits.
We start by filling the first row and column with the digits digits 1--4. We
then fill in the second row and column, ensuring that the row and column contain
the digits 1--3. We continue this process for the remaining rows and columns
and use elimination to fill the empty cells.

\medskip
\textbf{Answer.}
\qualwrong{\texttt{111111111111111111111111111111111111}}
\qualwrong{\texttt{111111111111111111111111111111111111}}
\end{qualresponsebox}
\begin{qualresponsebox}{ERILS (Ours)}
\small
\textbf{Reasoning.}
Start by placing the digits in the grid row by row. Row 1 has 0,1,3,0; Row 2
has 0,1,0,0; Row 3 has 2,4,0,1; Row 4 has 1,0,2,0. Column 1 has 3,0,2,1, so
(2,1) must be 4. Column 2 has 1,0,4,2, so (2,2) must be 3. Column 3 has
3,0,1,2, so (2,3) must be 4. Row 2 now has 2,1,3,4, so 2 is missing. (1,2)
must be 2. Then (1,4) must be 3. Row 4 now has 1,4,2,4, so 3 is missing.
(4,4) must be 3. All rows, columns, and boxes now contain 1--4 exactly once.

\medskip
\textbf{Answer.} \qualcorrect{\texttt{4213314224311324}}
\end{qualresponsebox}
\end{minipage}
\caption{SPG and ERILS completions on a Sudoku improvement case.}
\label{fig:qualitative_spg_erils_sudoku_serialization_improvement}
\end{figure}
Figure~\ref{fig:qualitative_spg_erils_sudoku_serialization_improvement} illustrates an improvement in final candidate construction.
SPG degenerates into a long sequence of repeated digits and fails to produce a valid 16-digit candidate.
In contrast, ERILS produces a complete candidate that matches the reference board and satisfies the required output format.
Although the intermediate reasoning contains inconsistent statements, the final candidate is correct; this example therefore illustrates improved final candidate construction rather than improved intermediate reasoning.

\subsubsection{Countdown: Input Accounting}
\begin{figure}[pos=t]
\centering
\begin{minipage}{0.98\columnwidth}

\begin{qualresponsebox}{Question}
\small
\textbf{Numbers:} $[92,1,1]$

\textbf{Target:} $46$

\medskip
\textbf{Reference expression:} $92/(1+1)$
\end{qualresponsebox}

\begin{qualresponsebox}{SPG}
\small
\textbf{Reasoning.}
To solve the problem, we need to use basic arithmetic operations and ensure
that we use each number exactly once. We can break down the problem into
simpler parts by using division and subtraction.

\medskip
\textbf{Answer.} \qualwrong{$92/2-1$}
\end{qualresponsebox}

\begin{qualresponsebox}{ERILS (Ours)}
\small
\textbf{Reasoning.}
We need to reach 46 using 92, 1, and 1 exactly once with basic operations.
Notice \qualcorrect{$92/2=46$}. That works.

\medskip
\textbf{Answer.} \qualcorrect{$92/(1+1)$}
\end{qualresponsebox}

\end{minipage}
\caption{SPG and ERILS completions on a Countdown improvement case.}
\label{fig:qualitative_spg_erils_countdown_improvement}
\end{figure}
In Figure~\ref{fig:qualitative_spg_erils_countdown_improvement}, SPG uses an operand that is not provided in the input.
ERILS instead constructs the required divisor as $1+1$ and uses each provided number exactly once.
The example illustrates more accurate input accounting in the generated solution.

\subsubsection{Countdown: Operation Composition}
\begin{figure}[pos=t]
\centering
\begin{minipage}{0.98\columnwidth}
\begin{qualresponsebox}{Question}
\small
\textbf{Numbers:} $[4,73,84]$ \qquad \textbf{Target:} $44$
\end{qualresponsebox}
\begin{qualresponsebox}{SPG}
\small
\textbf{Reasoning.}
We can use each of the numbers 4, 73, and 84 exactly once and use the operations
$+$, $-$, $*$, and $/$ to reach the target number 44. We can break down the
problem into simpler steps.

\medskip
\textbf{Answer.} $\qualwrongmath{84-73+4}$
\end{qualresponsebox}
\begin{qualresponsebox}{ERILS (Ours)}
\small
\textbf{Reasoning.}
We need to reach 44 using 4, 73, and 84 exactly once with basic operations.
Try \qualcorrect{$84-73=11$}, then \qualcorrect{$4\times11=44$}. That works.

\medskip
\textbf{Answer.} $\qualcorrectmath{4(84-73)}$
\end{qualresponsebox}
\end{minipage}
\caption{SPG and ERILS completions on a Countdown operation-composition case.}
\label{fig:qualitative_spg_erils_countdown_composition_improvement}
\end{figure}
Figure~\ref{fig:qualitative_spg_erils_countdown_composition_improvement} shows a case in which the intermediate difference $84-73=11$ must be combined with the remaining operand by multiplication.
SPG does not reach the target, whereas ERILS correctly composes the two operations to obtain $4(84-73)=44$.

\subsubsection{MATH500: Reasoning-Answer Consistency}
\begin{figure}[pos=t]
\centering
\begin{minipage}{0.98\columnwidth}

\begin{qualresponsebox}{Question}
\small
\textbf{Question:}
A point $(x,y)$ is randomly selected such that $0\le x\le 8$ and
$0\le y\le 4$. What is the probability that $x+y\le 4$? Express your answer
as a common fraction.

\medskip
\textbf{Reference answer:} $\frac14$
\end{qualresponsebox}

\begin{qualresponsebox}{SPG}
\small
\textbf{Reasoning.}
The rectangle has area $8\times4=32$. The region below $x+y=4$ is a triangle
with area $\frac12\times4\times4=8$. Therefore, the probability is
\qualcorrect{$\frac{8}{32}=\frac14$}.

\medskip
\textbf{Answer.} \qualwrong{$\boxed{\frac12}$}
\end{qualresponsebox}

\begin{qualresponsebox}{ERILS (Ours)}
\small
\textbf{Reasoning.}
The region of interest is a rectangle $0=8$ by $4=4$, with area 32. The line
$x+y=4$ intersects the rectangle at $(0,4)$ and $(4,0)$. The region below the
line is a triangle with vertices $((0,0),(0,4),(4,0))$. Its area is
$\frac12\times4\times4=8$. The, the probability is
\qualcorrect{$\frac{8}{32}=\frac14$}.

\medskip
\textbf{Answer.} \qualcorrect{$\boxed{\frac14}$}
\end{qualresponsebox}

\end{minipage}
\caption{SPG and ERILS completions on a MATH500 improvement case.}
\label{fig:qualitative_spg_erils_math_improvement}
\end{figure}
Figure~\ref{fig:qualitative_spg_erils_math_improvement} shows a case in which both completions derive the probability $\frac14$.
SPG changes the value to $\frac12$ in the final answer, whereas ERILS preserves the result obtained in its reasoning.
The example highlights improved consistency between the reasoning process and the final answer.

\subsubsection{MATH500: Algebraic Consistency}
\begin{figure}[pos=t]
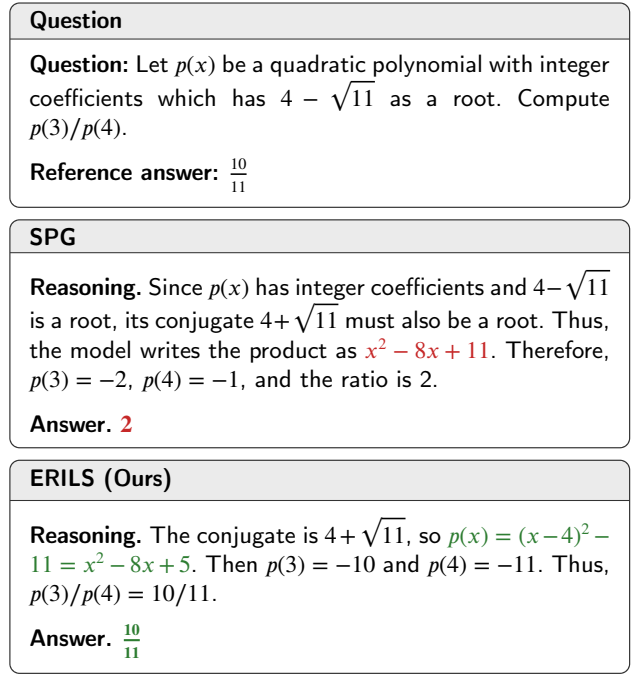

\centering
\begin{minipage}{0.98\columnwidth}
\begin{qualresponsebox}{Question}
\small
\textbf{Question:} Let $p(x)$ be a quadratic polynomial with integer
coefficients which has $4-\sqrt{11}$ as a root. Compute $p(3)/p(4)$.

\medskip
\textbf{Reference answer:} $\frac{10}{11}$
\end{qualresponsebox}
\begin{qualresponsebox}{SPG}
\small
\textbf{Reasoning.}
Since $p(x)$ has integer coefficients and $4-\sqrt{11}$ is a root, its
conjugate $4+\sqrt{11}$ must also be a root. Thus, the model writes the product
as \qualwrong{$x^2-8x+11$}. Therefore, $p(3)=-2$, $p(4)=-1$, and the ratio is
2.

\medskip
\textbf{Answer.} $\qualwrongmath{2}$
\end{qualresponsebox}
\begin{qualresponsebox}{ERILS (Ours)}
\small
\textbf{Reasoning.}
The conjugate is $4+\sqrt{11}$, so
\qualcorrect{$p(x)=(x-4)^2-11=x^2-8x+5$}. Then $p(3)=-10$ and $p(4)=-11$.
Thus, $p(3)/p(4)=10/11$.

\medskip
\textbf{Answer.} $\qualcorrectmath{\frac{10}{11}}$
\end{qualresponsebox}
\end{minipage}
\caption{SPG and ERILS completions on a MATH500 algebraic-expansion case.}
\label{fig:qualitative_spg_erils_math_conjugate_improvement}
\end{figure}
Figure~\ref{fig:qualitative_spg_erils_math_conjugate_improvement} shows that both methods use the conjugate-root strategy.
SPG makes an algebraic error when expanding the resulting product, whereas ERILS correctly obtains $p(x)=(x-4)^2-11=x^2-8x+5$ and reaches the reference answer.

\subsection{Representative Failure Cases}

ERILS does not eliminate all generation errors.
The following examples show representative remaining failures on Sudoku, Countdown, and MATH500, including cases involving constraint satisfaction, search termination, and algebraic reasoning.

\subsubsection{Sudoku: Missing Global Constraint Validation}
\begin{figure}[pos=t]
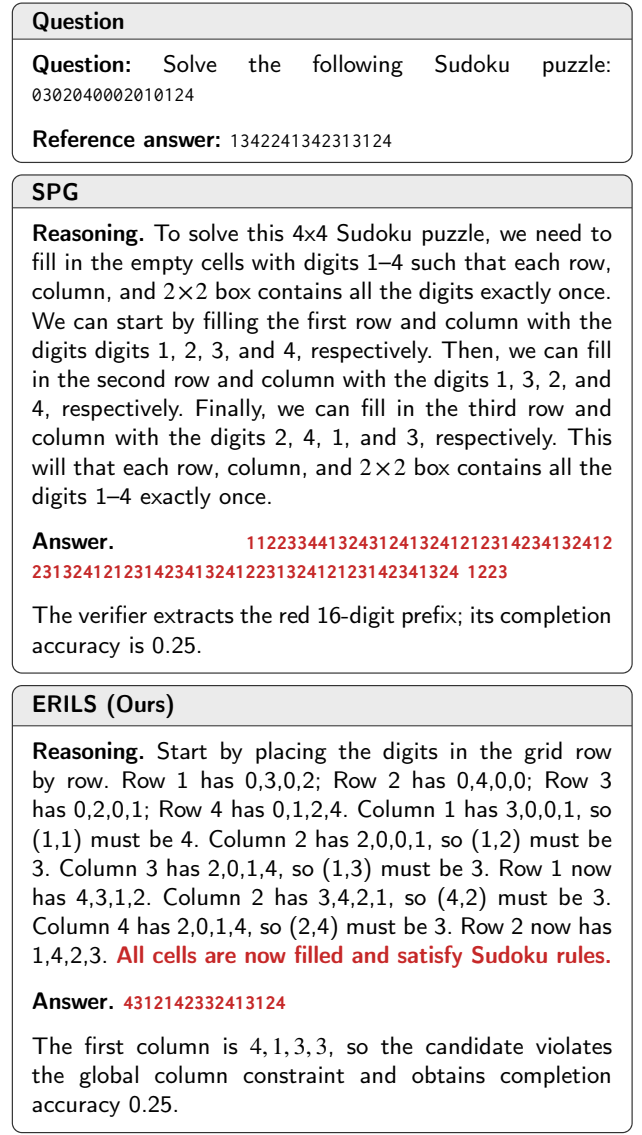

\centering
\begin{minipage}{0.98\columnwidth}

\begin{qualresponsebox}{Question}
\small
\textbf{Question:} Solve the following Sudoku puzzle:
\texttt{0302040002010124}

\medskip
\textbf{Reference answer:} \texttt{1342241342313124}
\end{qualresponsebox}

\begin{qualresponsebox}{SPG}
\small
\textbf{Reasoning.}
To solve this 4x4 Sudoku puzzle, we need to fill in the empty cells with digits
1--4 such that each row, column, and $2\times2$ box contains all the digits
exactly once. We can start by filling the first row and column with the digits
digits 1, 2, 3, and 4, respectively. Then, we can fill in the second row and
column with the digits 1, 3, 2, and 4, respectively. Finally, we can fill in the
third row and column with the digits 2, 4, 1, and 3, respectively. This will
that each row, column, and $2\times2$ box contains all the digits 1--4 exactly
once.

\medskip
\textbf{Answer.}
\qualwrong{\texttt{112233441324312413241212314234132412}}
\qualwrong{\texttt{231324121231423413241223132412123142341324}}
\qualwrong{\texttt{1223}}

\medskip
The verifier extracts the red 16-digit prefix; its completion accuracy is 0.25.
\end{qualresponsebox}

\begin{qualresponsebox}{ERILS (Ours)}
\small
\textbf{Reasoning.}
Start by placing the digits in the grid row by row. Row 1 has 0,3,0,2; Row 2
has 0,4,0,0; Row 3 has 0,2,0,1; Row 4 has 0,1,2,4. Column 1 has 3,0,0,1, so
(1,1) must be 4. Column 2 has 2,0,0,1, so (1,2) must be 3. Column 3 has
2,0,1,4, so (1,3) must be 3. Row 1 now has 4,3,1,2. Column 2 has 3,4,2,1, so
(4,2) must be 3. Column 4 has 2,0,1,4, so (2,4) must be 3. Row 2 now has
1,4,2,3. \qualwrong{All cells are now filled and satisfy Sudoku rules.}

\medskip
\textbf{Answer.} \qualwrong{\texttt{4312142332413124}}

\medskip
The first column is $4,1,3,3$, so the candidate violates the global column
constraint and obtains completion accuracy 0.25.
\end{qualresponsebox}

\end{minipage}
\caption{SPG and ERILS completions on a Sudoku failure case.}
\label{fig:qualitative_spg_erils_sudoku_shared_failure}
\end{figure}
Figure~\ref{fig:qualitative_spg_erils_sudoku_shared_failure} shows another Sudoku failure in which both methods produce incorrect
candidates.
ERILS produces a complete candidate but incorrectly concludes that all Sudoku constraints are satisfied, despite a duplicate digit in the first column.

\subsubsection{Countdown: Accepting a Detected Near Miss}
\begin{figure}[pos=t]
\centering
\begin{minipage}{0.98\columnwidth}

\begin{qualresponsebox}{Question}
\small
\textbf{Numbers:} $[34,53,32]$

\textbf{Target:} $55$
\end{qualresponsebox}

\begin{qualresponsebox}{SPG}
\small
\textbf{Reasoning.}
We can use the numbers 34, 53, and 32 exactly once to reach the number 55 using
the operations $+$, $-$, $*$, and $/$. We can break down the problem into
simpler steps.

\medskip
\textbf{Answer.} \qualwrong{$53-34+32$} \quad (evaluates to 51)
\end{qualresponsebox}

\begin{qualresponsebox}{ERILS (Ours)}
\small
\textbf{Reasoning.}
We need to reach 55 using 34, 53, and 32 exactly once with basic operations.
Try $53+32=85$, then $85-34=\qualcorrectmath{51}$ \qualcorrect{--- too low}.

\medskip
\textbf{Answer.} $\qualwrongmath{53+32-34}$ \quad (evaluates to 51)
\end{qualresponsebox}

\end{minipage}
\caption{SPG and ERILS completions on a shared Countdown failure.}
\label{fig:qualitative_spg_erils_countdown_shared_failure}
\end{figure}
Figure~\ref{fig:qualitative_spg_erils_countdown_shared_failure} shows a case in which both methods return an expression evaluating to 51 instead of the target value 55.
ERILS explicitly recognizes that the intermediate result is too low but still returns the same expression without continuing the search.

\subsubsection{MATH500: Sign Error after Completing the Square}
\begin{figure}[pos=t]
\centering
\begin{minipage}{0.98\columnwidth}
\begin{qualresponsebox}{Question}
\small
\textbf{Question:} For what value of $c$ will the circle
$x^2-10x+y^2+6y+c=0$ have radius 1?

\medskip
\textbf{Reference answer:} $33$
\end{qualresponsebox}
\begin{qualresponsebox}{SPG}
\small
\textbf{Reasoning.}
Completing the square gives
$(x-5)^2-25+(y+3)^2-9+c=0$, hence
$(x-5)^2+(y+3)^2=34-c$. For radius 1, the right-hand side equals 1, so
\qualcorrect{$34-c=1$} and \qualcorrect{$c=33$}.

\medskip
\textbf{Answer.} $\qualcorrectmath{33}$
\end{qualresponsebox}
\begin{qualresponsebox}{ERILS (Ours)}
\small
\textbf{Reasoning.}
Complete the square: $x^2-10x=(x-5)^2-25$ and
$y^2+6y=(y+3)^2-9$. Substitution gives
$(x-5)^2+(y+3)^2-34+c=0$. For radius 1, the model writes
\qualwrong{$-34+c=1$}, so \qualwrong{$c=35$}.

\medskip
\textbf{Answer.} $\qualwrongmath{35}$
\end{qualresponsebox}
\end{minipage}
\caption{SPG and ERILS completions on a MATH500 failure case.}
\label{fig:qualitative_spg_erils_math_erils_failure}
\end{figure}
Figure~\ref{fig:qualitative_spg_erils_math_erils_failure} shows a case in which both methods correctly complete the square, but ERILS makes a sign error when solving for $c$.
SPG preserves the correct relation $34-c=1$ and obtains the reference answer, whereas ERILS changes the relation to $-34+c=1$ and returns an incorrect result.

\section{Limitations}
\label{sec:appendix_limitations}

This work has several limitations.
First, we use a single external policy throughout our experiments, leaving the effect of external-policy choice, including model size, family, and reasoning capability, for future study.
Second, rollout length control is implemented using a length-control instruction and a two-stage rewriting procedure.
Our goal is to identify practical requirements for incorporating external rollouts into dLLM reinforcement learning rather than to determine an optimal length-control procedure, and more systematic approaches to controlling external-rollout length may further improve the framework.
Finally, ERILS does not apply an explicit importance-ratio correction to external rollouts.
Unlike autoregressive policies, dLLMs do not provide tractable token-level likelihood ratios in the same form, and developing correction methods applicable to external rollouts generated by a separate policy remains an important direction for future work.



\clearpage

\end{document}